\pdfoutput=1

\documentclass[11pt]{article}

\usepackage[final]{acl}
\usepackage{times}
\usepackage{latexsym}

\usepackage[T1]{fontenc}

\usepackage[utf8]{inputenc}

\usepackage{microtype}

\usepackage{inconsolata}

\usepackage{graphicx}

\usepackage{microtype}
\usepackage{hyperref}
\usepackage{url}
\usepackage{booktabs}
\usepackage{graphicx}
\usepackage{multirow}
\usepackage{array}
\usepackage{subcaption}
\usepackage{longtable}
\usepackage{marvosym}
\usepackage{amsmath,amsfonts,bm}
\usepackage{subcaption}

\newlength{\twosubht}
\newsavebox{\twosubbox}
\title{Uncovering Latent Arguments in Social Media Messaging by Employing LLMs-in-the-Loop Strategy}

\author{Tunazzina Islam\\
  Department of Computer Science \\
  Purdue University, West Lafayette,\\
  Indiana-47907, USA\\
  \texttt{islam32@purdue.edu} \\\And
  Dan Goldwasser\\
  Department of Computer Science \\
  Purdue University, West Lafayette,\\
  Indiana-47907, USA\\
  \texttt{dgoldwas@purdue.edu} \\}

\begin{document}
\maketitle
\begin{abstract}
The widespread use of social media has led to a surge in popularity for automated methods of analyzing public opinion. Supervised methods are adept at text categorization, yet the dynamic nature of social media discussions poses a continual challenge for these techniques due to the constant shifting of the focus. On the other hand, traditional unsupervised methods for extracting themes from public discourse, such as topic modeling, often reveal overarching patterns that might not capture specific nuances.  
Consequently, a significant portion of research into social media discourse still depends on labor-intensive manual coding techniques and a human-in-the-loop approach, which are both time-consuming and costly. 
In this work, we study the problem of discovering arguments associated with a specific theme. We propose a generic \textbf{LLMs-in-the-Loop} strategy that leverages the advanced capabilities of large language models (LLMs) to extract latent arguments from social media messaging. 
To demonstrate our approach, we apply our framework to contentious topics. We use two publicly available datasets: (1) the climate campaigns dataset of $14k$ Facebook ads with $25$ themes and (2) the COVID-19 vaccine campaigns dataset of $9k$ Facebook ads with $14$ themes. Additionally, we design a downstream task as stance prediction by leveraging talking points in climate debates. Furthermore, we analyze demographic targeting and the adaptation of messaging based on real-world events.

\end{abstract}

\section{Introduction}
Public opinion is crucial for ensuring responsive governance, policy alignment with public interests, and societal harmony by facilitating a feedback loop for continuous policy refinement \citep{glynn2008public,price1988public}.
The advent of social media has transformed the landscape of public discourse and opinion formation, enabling the rapid exchange of information and ideas, facilitating instant communication, and promoting a participatory approach to addressing societal issues \citep{mcgregor2019social,xiong2014opinion,murphy2014social}.
This shift has led to an increased focus on automatically analyzing public opinion on social media platforms \citep{liu2022sentiment,pacheco2022holistic,yousefinaghani2021analysis,han2020using,islam2020does,liang2013opinion,maynard2012challenges,sobkowicz2012opinion}.
 In response to this, argument mining has emerged as a critical technique that automatically extracts the reasons, claims, and talking points (arguments)\footnote{We will use arguments and talking points interchangeably in this paper.}, shedding light on how and why specific opinions are formed \citep{wawrzuta2021arguments,sowa2021covid,skeppstedt2018vaccine,habernal2018argument,stab2014identifying,kim2006automatic}. In these cases, the variables of interest are well
defined, and substantial efforts are invested in developing manually annotated resources, thereby allowing the problems to be structured as supervised learning tasks.

Unsupervised text analysis methods like topic modeling (Latent Dirichlet Allocation (LDA) \citep{blei2003latent}, non-negative matrix factorization (NMF) \citep{lee1999learning}) can discover topics within data without requiring pre-labeled datasets. However, they fall short of extracting meaningful themes beneath those topics. Therefore, a non-trivial number of recent studies on social media discourse rely on manual and qualitative coding methods \citep{hagen2022role,NGUYEN2021100922,del2020online}.

In recent years, efforts have been made to delve into the subtleties of topics in social media content by discovering themes \citep{islam2024uncoveringthm,pacheco2023interactive,islam2023analysis,islam2022understanding,pacheco2022holistic}.
Yet, to fully comprehend why an opinion is formed, it's crucial to identify the specific arguments within these broader themes. For example,
under the `Climate Change' topic, a theme might be `Alternative Energy', and under this theme, there might be conflicting arguments, such as Argument 1: \textit{``Alternative energy will create more jobs.”} versus
Argument 2: \textit{``Alternative energy will take away our jobs.”}. 
Identifying arguments within a theme reveals the granular nuances.

\citet{pacheco2022interactively,pacheco2022holistic} has developed a human-in-the-loop approach for discovering arguments by balancing unsupervised text analysis techniques and manual coding. While this method enhances accuracy, it suffers from scalability issues, is time-consuming, and demands high resources. In this paper, we introduce an LLMs-in-the-Loop approach that integrates the capabilities of Large Language Models (LLMs) \citep{brown2020language} and incorporates them into an efficient algorithmic framework to uncover latent arguments under a theme in social media messaging.

\begin{figure}[t]
\includegraphics[width=\columnwidth]{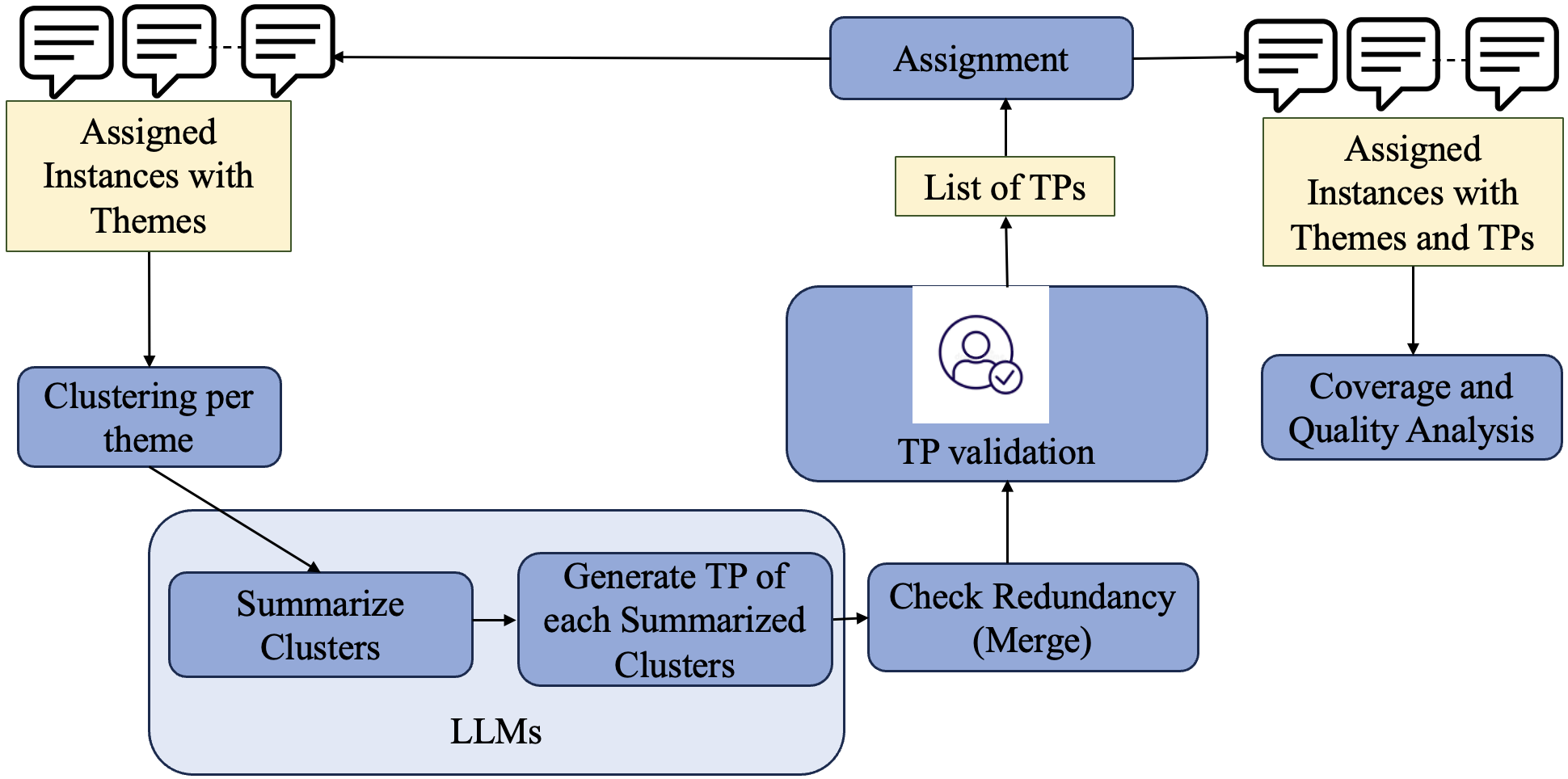}
\caption{LLMs-in-the-Loop framework. TP: Talking point.} \label{fig1}
\end{figure}
We define large textual collections as repositories of textual instances (e.g., ads, tweets, posts, documents) where each instance is associated with a theme (the set of themes derived from previous studies \citep{islam2024uncoveringthm,islam2023analysis,pacheco2023interactive,islam2022understanding,pacheco2022holistic}).
An overview of our framework is illustrated in Fig. \ref{fig1}.
Our framework provides a way of clustering talking points motivated by themes. We have assigned instances with themes. For each theme cluster, we cluster the associated instances using a clustering algorithm. 
We characterize the sub-clusters by performing zero-shot multi-document summarization of the top-K instances assigned to each sub-cluster. Then, we prompt LLMs in a zero-shot manner to generate the talking point advocating in the sub-cluster summary in the context of the given theme.
For redundancy check, we assess the similarity between talking point pairs and merge overlapping talking points based on threshold. 
We have an optional human evaluation phase for newly generated talking points. Finally, the assignment phase maps instances to their most likely talking points.

Our framework is designed to address three main challenges: 1) Given a large collection of instances with a known space of high-level themes, how can we effectively
identify a set of repeating arguments, 2) how can we refine those arguments, and 3) how can we map instances to corresponding arguments efficiently.
To demonstrate our method, we look at the task of
characterizing social media messaging on contentious topics.
Various interest groups, including politicians, advertisers, and stakeholders, have used the highly distributed social media landscape to microtarget audiences \citep{hersh2015,barbu2014advertising}. While effective in enhancing content relevance, microtargeting also presents risks such as manipulating user behavior \citep{ribeiro2019microtargeting,etudo2019facebook,kruikemeier2016political}, creating echo chambers \citep{garimella2018political,lima2018inside,del2016spreading,quattrociocchi2016echo,jamieson2008echo}, and fostering polarization \citep{jiang2020political,zuiderveen2018online,bostrom2013targeting}. 
%

In this paper, we examine two distinct case
studies: the \textbf{climate campaigns}
and the \textbf{COVID-19 vaccine campaigns} in the United States. 
In each case, the qualitative researchers
apply different theories to identify themes. For climate campaigns, the identification of themes is based on 
the energy industry and climate change-related stances \citep{islam2023analysis}. For the COVID-19 vaccine campaigns \citep{islam2022understanding}, the theme discovery is grounded using Moral Foundation Theory (MFT) \citep{haidt2007morality,haidt2004intuitive}. 
These case studies are selected for their relevance to the computational social science (CSS) community, illustrating the complexities of targeted messaging in a digital age. 
From a machine learning viewpoint, these case studies represent distinct challenges because of the data, context, and target themes.

Our experiments show that our framework can
be used to uncover a set of arguments that cover a large portion of the discussion about the climate and COVID-19 vaccine campaigns on Facebook and the resulting mapping
from ads to arguments is fairly accurate with respect to human judgments. Moreover, we introduce a downstream task of stance classification that leverages uncovered talking points, intuitively designed to elucidate the associations between stances and the talking points employed by advertisers in contentious debates. Additionally, we provide an in-depth analysis of how messaging is tailored to specific demographics and how these talking points evolve in response to real-world events.
%
\section{Related Work}
Previous works for understanding messaging on social media \citep{islam2023weakly,islam2023analysis,islam2022understanding,capozzi2021clandestino} used a predefined set of labels,
themes, and arguments to analyze messaging. These were
fixed and established based on existing topics or theoretical frameworks. In this work, we focus on gaining a deeper understanding of messaging, especially at the argument level.

Argument mining involves identifying and extracting argument components such as claims, evidence, and conclusions and understanding how they are logically connected to form coherent arguments. Several works have been done on analyzing arguments on social media \citep{islam2023analysis,pacheco2022interactively,pacheco2022holistic,islam2022understanding,wawrzuta2021arguments,bhatti2021argument,malagoli2021look,kotelnikov2022ruarg,freire2019characterization,skeppstedt2018vaccine,torsi2018annotating,dusmanu2017argument}. 
\citet{bhatti2021argument} formulated argument mining on Twitter as a text classification task to identify tweets serving as premises for hashtags that represent claims to reveal prominent arguments for and against funding Planned Parenthood. \citet{wawrzuta2021arguments} studied the arguments made by Twitter users in Poland when discussing the COVID-19 vaccine. \citet{pacheco2022interactively,pacheco2022holistic} proposed interactive human-in-the-loop protocol to analyze COVID-19 vaccine debate in Twitter.
\citet{miller2016audience} characterized the arguments supporting the oil and gas industries. \citet{islam2023analysis} manually developed relevant arguments for climate change and energy industry-related stances. In this paper, instead of using manual coding and human feedback to discover arguments, we rely on a machine-in-the-loop approach.

Recently, large language models (LLMs) have achieved
promising progress in learning from prompts via in-context learning (ICL) \citep{chowdhery2023palm,kojima2022large,le2022bloom,brown2020language}.
Several studies indicate that LLMs exhibit better performance in tasks traditionally completed by humans \citep{gilardi2023chatgpt,de2023can,dai2023llm,chiang2023can,ziems2024can}, highlighting a potential to leverage LLMs effectively in our task. A recent approach has introduced the concept of an LLM-in-the-loop for thematic analysis \citep{dai2023llm}. In contrast, our framework employs LLMs-in-the-Loop to uncover the latent arguments in messaging.
At a high level, our work connects with prior work on interactive clustering \citep{pacheco2023interactive,pacheco2022holistic,lund2017tandem,hu2014interactive,bernstein2010eddi}, but instead of using human feedback to shape the emergent clusters, we rely
on LLM inference.
%
%
\section{LLMs-in-the-Loop Framework}
\label{llm}
We propose an iterative LLMs-in-the-Loop framework that combines Natural Language Processing (NLP) techniques and inference capabilities of LLMs to automate the process of discovering latent arguments within themes. Our framework enhances the ability to analyze vast amounts of social media content by identifying nuanced arguments within predefined themes. This section outlines the methodological steps, including clustering instances by themes, identifying arguments, refining those arguments, and mapping instances to corresponding arguments.
%
\subsection{Theme-Specific Clustering}
Initially, the framework categorizes textual instances into clusters based on their associated themes. These instances can range from social media posts to comprehensive documents. We divide each theme-based cluster into sub-clusters using a clustering algorithm. This allows for a more granular analysis of the thematic content, revealing the nuances of arguments present within each theme. We
embed the instances using Sentence BERT \citep{reimers2019sentence}. The embedded instances are clustered using K-means \citep{Jin2010}. To determine the optimal value of $k$ in k-Means, we follow both the Elbow method and the Silhouette method. 
\subsection{Summarizing Sub-clusters}
To articulate the arguments found within each sub-cluster, we employ zero-shot multi-document summarization using GPT-4 \citep{achiam2023gpt} on the top-k instances. The prompts are engineered to generate short theme-specific summaries in a zero-shot setting. The five closest instances to each centroid are used in the prompt to generate the summaries. This summarization process highlights the key points and arguments without the need for pre-labeled data, showcasing the framework's unsupervised capabilities.
\subsection{Generating and Refining Arguments}
Subsequently, each sub-cluster summaries serve as a prompt for LLMs in a zero-shot manner to generate specific talking points advocating for the arguments implied in the summaries. This approach ensures that the extracted talking points are relevant and coherent within the context of their respective themes.
%
\subsubsection{Argument Redundancy Check}
To refine the generated arguments, we implement a redundancy check to identify and merge similar talking points. This process involves assessing the similarity between pairs of talking points and consolidating those that exceed a predefined similarity threshold. We embed the talking points using Sentence BERT and compute the cosine similarity score between the embedded talking point pairs. Based on a threshold value ($\geq 0.70$), we decide to merge two arguments. Here are the two examples of talking points where we decide to merge them based on the threshold value- TP1: \textit{"The Build Back Better Act is crucial for economic growth, job creation, and addressing the climate crisis."};
TP2: \textit{"Legislative support for the Build Back Better Act, emphasizing its benefits for clean energy jobs and climate action."}

%
\subsection {Human Evaluation}
An optional human evaluation phase is incorporated to ensure the quality and relevance of the generated talking points. This phase allows for the verification of generated and refined talking points, ensuring their alignment with human understanding and interpretation. The annotators are asked to verify if the generated talking points and the merging decision are correct or not. They are
asked to provide a score of $1$ if the talking points are correct and a score of $0$ if they are not.
\subsection {Mapping Instances to Arguments}
In this stage of our framework, we use a simple distance-based approach for mapping. To measure the closeness between an instance and an argument, we compute the cosine distance between the instance and all arguments for each theme and take the minimum distance score among them.
We embed the arguments and instances using Sentence BERT. By thresholding, we determine if the instance can
be mapped to an argument. 
\begin{table*}[t]
\begin{center}
\begin{tabular}{l|lllllll}
\toprule
{\bf Case Study}  &  {\bf Iter.}  & {\bf \# Args} & {\bf Coverage} & {\bf $\leq Q_1$} & {\bf $\leq Q_2$} & {\bf $\leq Q_3$} & {\bf All}\\
\midrule
\multirow{2}{*}{Climate} & 1 &  113   & 37.38\% & 76.00\% & 70.67\% & 58.67\% & 57.33\% \\
& 2   &  213  & 44.40\%  & 88.00\% & 74.67\% & 70.67\% & 64.00\%  \\
\midrule
\midrule
\multirow{2}{*}{COVID-19} & 1 &  47   & 36.18\% & 78.57\% & 61.90\% & 61.90\% & 52.38\% \\
& 2   &  78  & 40.47\%  & 82.93\% & 73.81\% & 64.29\% & 57.14\%  \\
\bottomrule
\end{tabular}
\end{center}
\caption{Coverage and mapping quality w.r.t. Human Judgments. }
\label{result}
\end{table*}
\section{Case Studies}
We explore two case studies involving discussions
on social media: (1) climate campaigns and (2) COVID-19 vaccine campaigns. For climate campaigns, we work on the corpus of $14k$ ads released by \citet{islam2023analysis}. The dataset focuses on climate-related English ads on Facebook based on the US from January $2021$  to January $2022$.  All ads in this
corpus contain predictions for stance (e.g.,
pro-energy, clean-energy) and theme (e.g., support climate policy.) \citep{islam2024uncoveringthm}.
For COVID-19 vaccine campaigns, we use the corpus of $9k$ ads released by \citet{islam2022understanding} focusing on COVID-19 vaccine-related English ads on Facebook based on US from December $2020$ to January $2022$. All ads in this corpus contain predictions for moral foundation (e.g., care/harm) \citep{haidt2007morality} and theme (e.g., vaccine equity.) \citep{islam2022understanding}. 
For each ad of both corpora, there are the following attributes: ad ID, title, ad description, ad body, funding entity, spend, impressions, distribution over impressions broken down by gender (male, female, unknown), age (7 groups), and location down to states in the USA. Derived themes are shown in Table \ref{tab1} in App. \ref{thm}. Data statistics regarding the number of instances in each theme for climate (Fig. \ref{stat_climate}) and COVID-19 vaccine (Fig. \ref{stat_covid}) campaign are shown in Fig. \ref{stat} in App. \ref{data_stat}. 
Additional details about the dataset can be found in the original publication.  

Our main goal is to use our LLMs-in-the-Loop framework to identify prominent arguments in the corpora described earlier.
At the beginning, we have ads and their associated themes. Those themes are derived by \citet{islam2024uncoveringthm,islam2022understanding}.
At the first iteration, we follow the step introduced in Section \ref{llm} to obtain the list of talking points under each theme and the mapping $ads \rightarrow arguments$. Hyperparameters are detailed in App. \ref{hyper}. After iteration $1$, we identify $113$ arguments for Climate campaigns and $47$ arguments for COVID-19 vaccine campaigns (Table \ref{result}).  In $2^{nd}$ iteration, we repeat this process for the rest of the ads remaining unassigned from iteration $1$ (by applying a threshold of $< 0.5$), resulting in the discovery of $100$ and $31$ additional arguments for climate and COVID-19 vaccine campaigns respectively (Table \ref{result}). An example of a resulting set of arguments under the `Patriotism' (one of the themes from climate campaigns) and `VaccineEquity' (one of the themes from COVID-19 vaccine campaigns) themes after two rounds of iterations of our LLMs-in-the-Loop are
shown in Table \ref{args} in App. \ref{arg_thm}. 
We show the number of sub-clusters identified under each theme for both iterations of the climate (Table \ref{sub_stat_climate}) and COVID-19 (Table \ref{sub_stat_covid}) case studies are provided in App. \ref{data_stat}.
The complete set of final arguments is provided in the App. \ref{arg_final} in Table \ref{arg_climate} and Table \ref{arg_covid}.
\begin{figure}[h]
\includegraphics[width=\columnwidth]{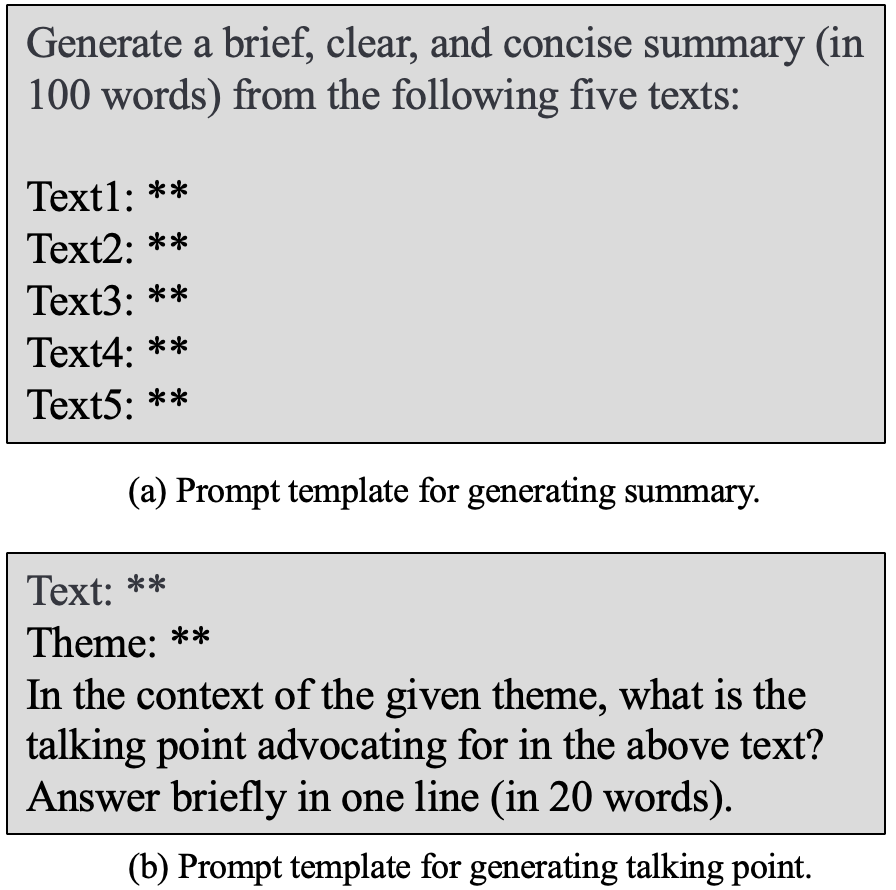}
\caption{Prompt templates (shown as zero-shot).} 
\label{prom}

\end{figure}
\subsection{Prompting}
\label{pt}
\begin{figure}[h]
\includegraphics[width=\columnwidth]{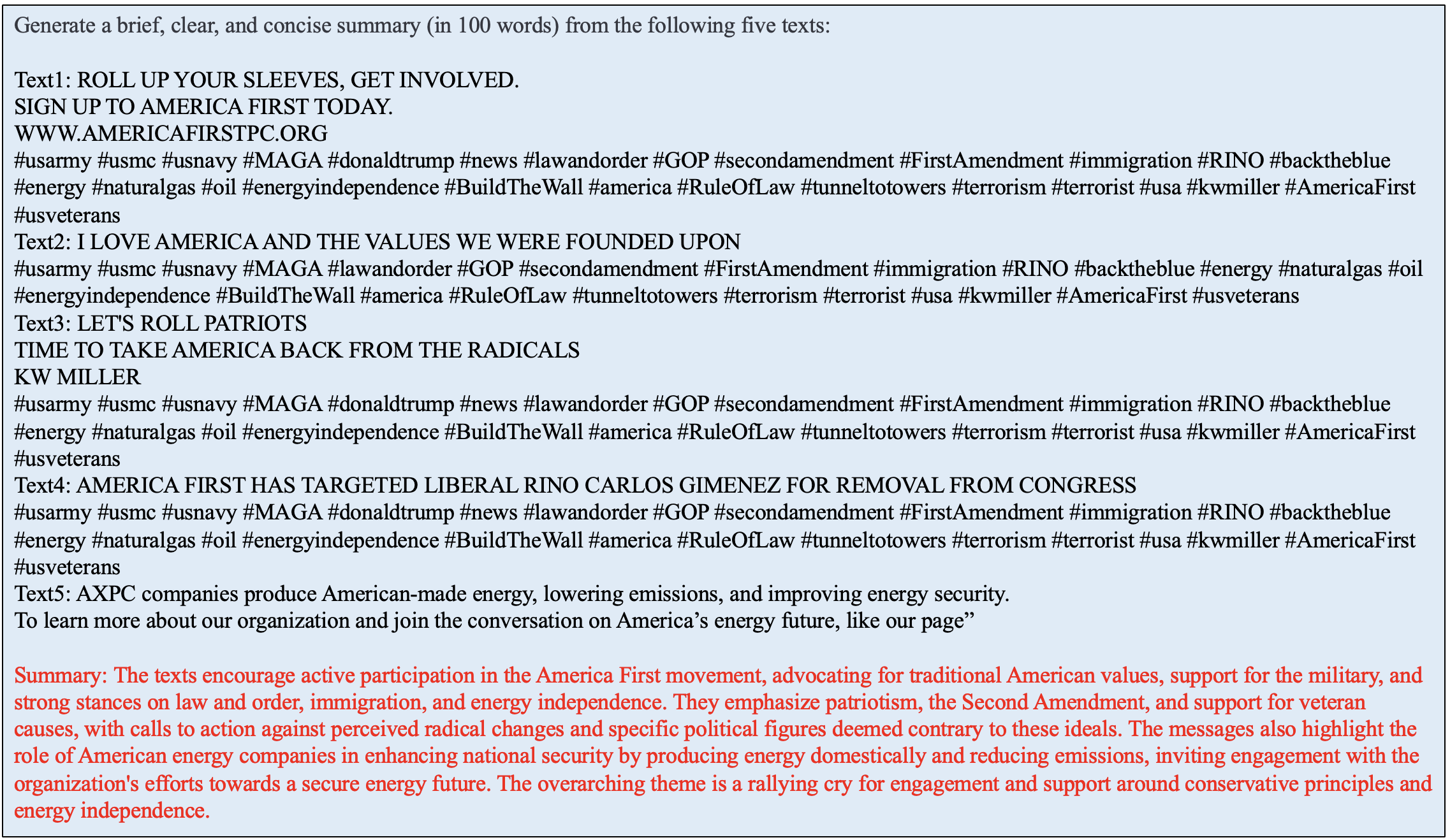}
\caption{Prompt example of summarizing top-5 instances under \textit{patriotism} theme of climate campaign dataset. The black colored segment is the input prompt and the red colored segment is the generated output by the LLMs.} 
\label{prompt_ex}

\end{figure}

\begin{figure}[h]
\includegraphics[width=\columnwidth]{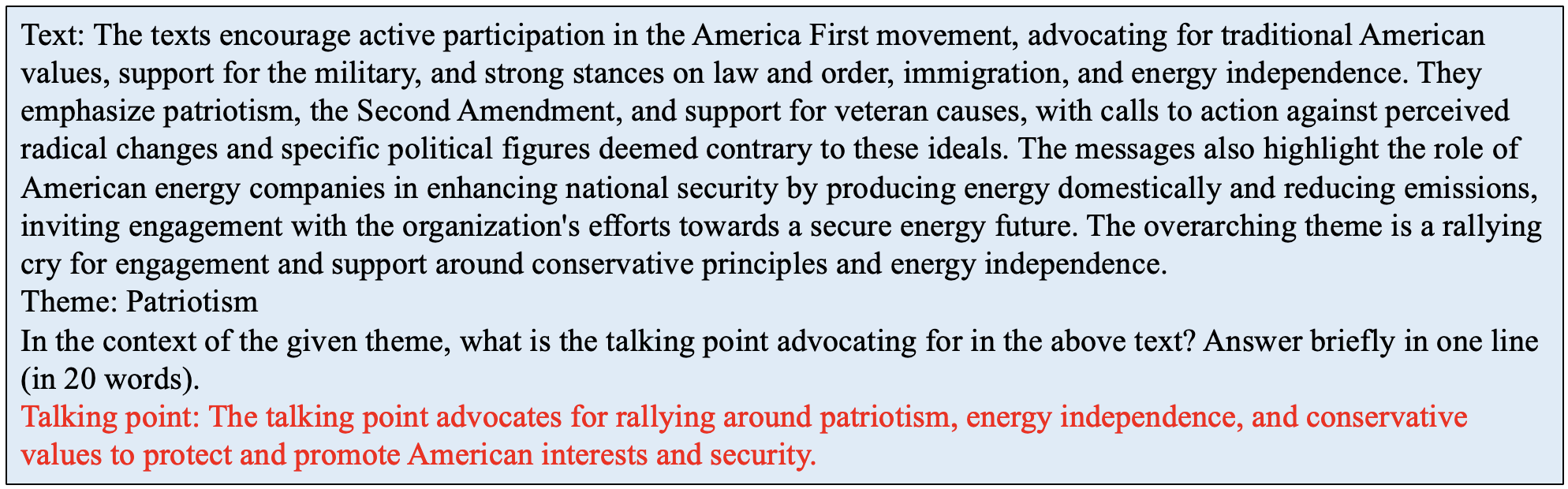}
\caption{Prompt example of generating talking point from a summary of top-5 instances under \textit{patriotism} theme of climate campaign dataset. The black colored segment is the input prompt and the red colored segment is the generated output by the LLMs.} 
\label{prompt_ex_tp}

\end{figure}

In this section, we show the prompt templates used in our work to generate a sub-cluster summary (Fig. \ref{prom} (a)) and a talking point from each summarized sub-cluster (Fig. \ref{prom} (b)). A concrete example of a prompt for summarizing top-5 instances under \textit{patriotism} theme of climate campaign and the generated summary are shown in Fig. \ref{prompt_ex}. A prompt example of generating a talking point from a summary of top-5 instances under \textit{patriotism} theme of the climate campaign dataset is illustrated in Fig. \ref{prompt_ex_tp}. 
\subsection{Evaluation}

To evaluate the performance of our ad-to-argument mapping, we sort the ads according to their semantic distance from their assigned arguments. We then compute the three quartiles and sample a set of $12$ ads per theme, such that $3$ ads are randomly sampled from each quartile. Subsequently, we manually annotate whether the mapping is correct or not (detailed in App. \ref{human}). This process results in $300$ ads in the $1^{st}$ iteration and another $300$ ads from the $2^{nd}$ iteration from the climate case study. On the other hand, from the COVID-19 vaccine case study, we obtain $168$ ads in the $1^{st}$ iteration and another $168$ ads from the $2^{nd}$ iteration.

To assess the performance across varying degrees of semantic closeness to the argument embeddings, we conduct evaluations within each quartile. Results for the first quartile (Q1) correspond to the $25\%$ closest examples. For the second quartile (Q2), they correspond to the $50\%$ closest examples,
and for the third quartile (Q3), to the $75\%$ closest examples. Intuitively, we expect better performance in the lower distance between the ad and the argument. Results are outlined in Table \ref{result}. 
We notice that we obtain higher macro average F1-scores for ads that are the closest to the arguments in the embedding space. In addition to the macro average F1-score, we also look at the percentage of ads that are covered by the set of arguments uncovered by our LLMs-in-the-Loop framework after each iteration. 
We do not enforce the idea that all ads need to be
mapped to arguments, and therefore, some ads remain unassigned. There is an improvement in performance both in coverage and mapping quality after subsequent iterations as we increase both the number of arguments and the number of
ads mapped (Table \ref{result}).







\begin{figure*}[t]
\centering
\begin{subfigure}{0.25\textwidth}
  \centering
  \includegraphics[width=\textwidth]{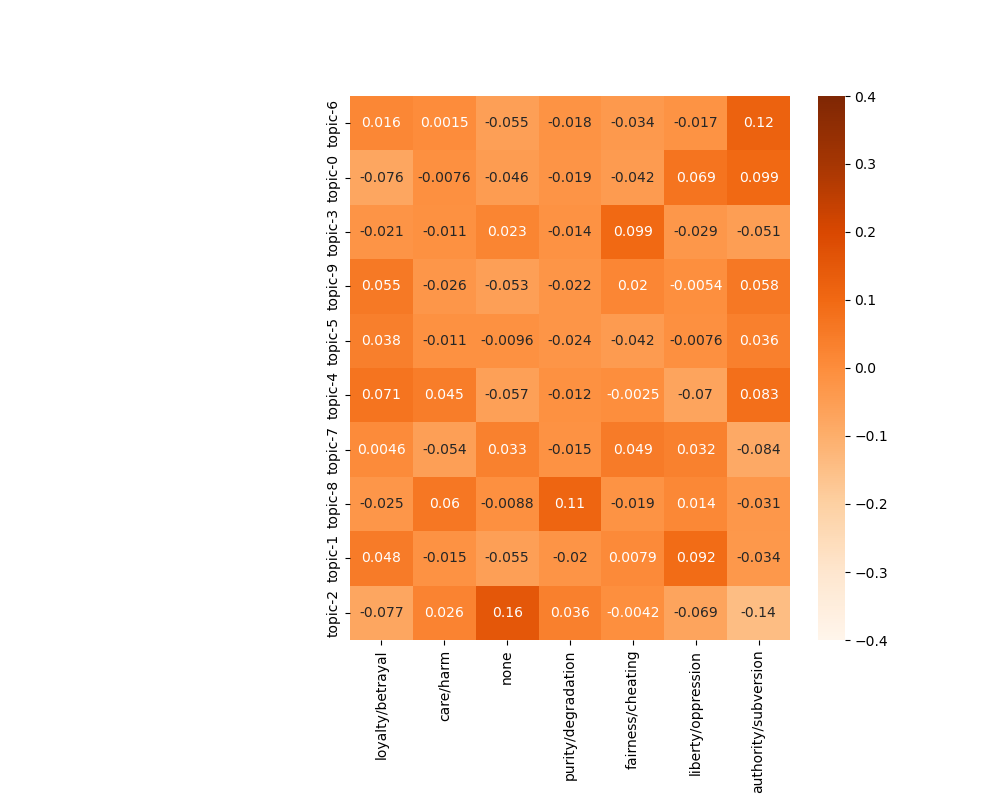}
  \caption{\textbf{Baseline}: 10 LDA Topics.}\label{fig:lda_1_covid}
\end{subfigure}%
\begin{subfigure}{0.25\textwidth}
  \centering
  \includegraphics[width=\textwidth]{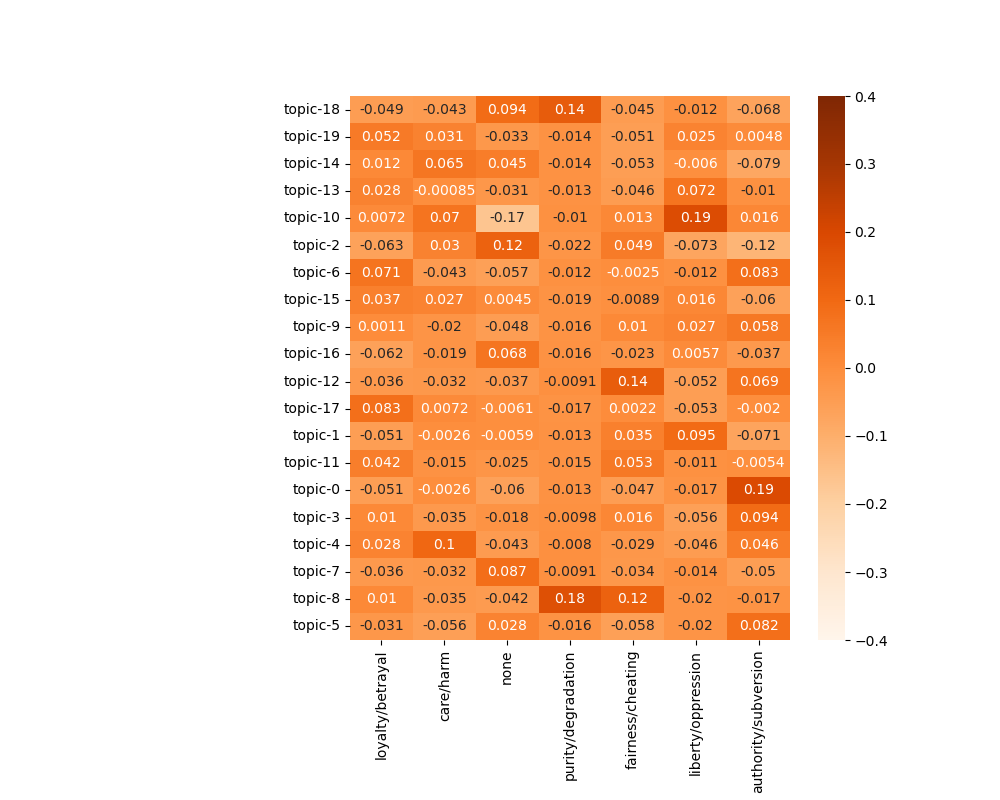}
  \caption{\textbf{Baseline}: 20 LDA Topics.}\label{fig:lda_2_covid}
\end{subfigure}%
\begin{subfigure}{0.5\textwidth}
  \centering
  \includegraphics[width=\textwidth]{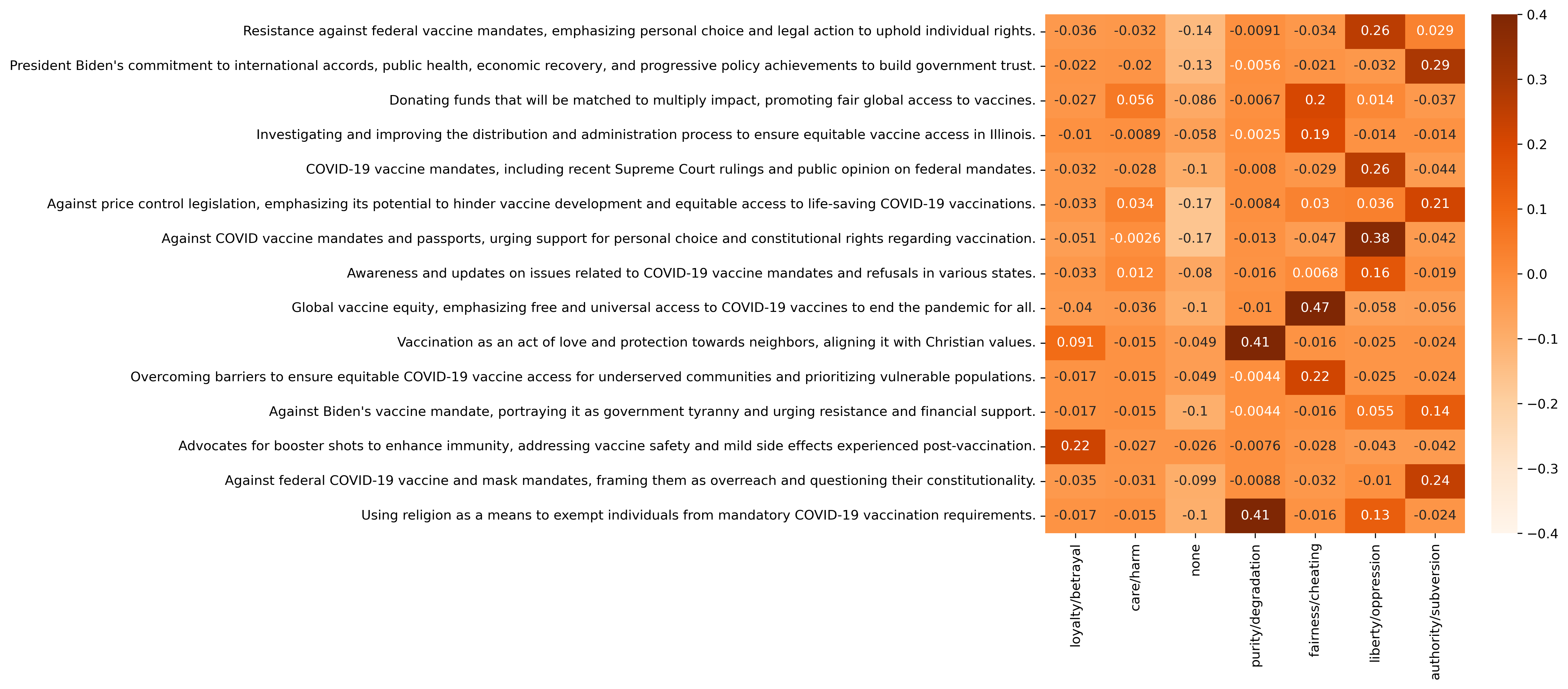}
  \caption{\textbf{Ours}: After $2^{nd}$ round of iteration (COVID-19).}\label{fig:ours_covid}
\end{subfigure}
\begin{subfigure}{0.25\textwidth}
  \centering
  \includegraphics[width=\textwidth]{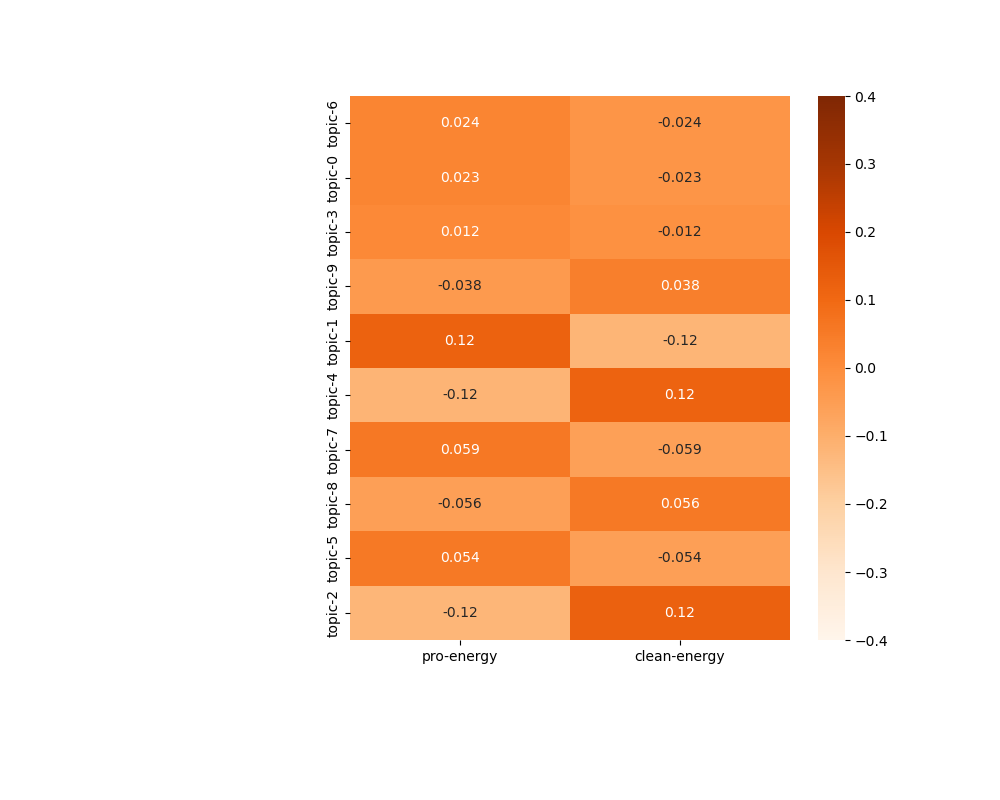}
  \caption{\textbf{Baseline}: 10 LDA Topics.}\label{fig:lda_1_climate}
\end{subfigure}%
\begin{subfigure}{0.25\textwidth}
  \centering
  \includegraphics[width=\textwidth]{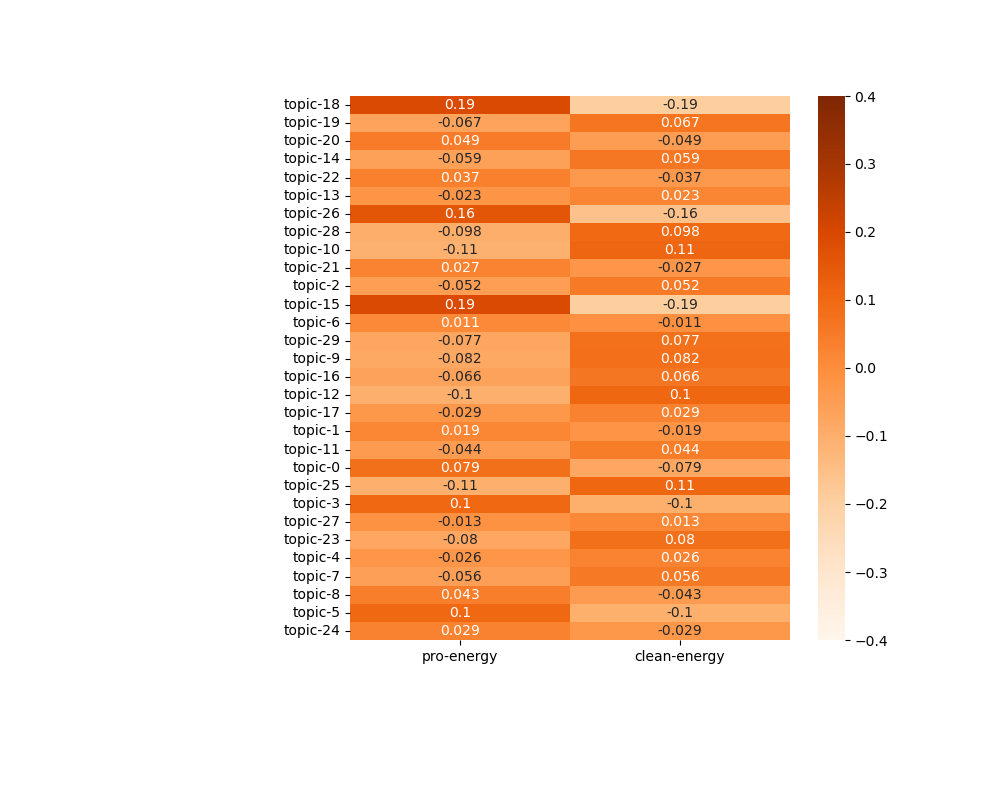}
  \caption{\textbf{Baseline}: 30 LDA Topics.}\label{fig:lda_3_climate}
\end{subfigure}%
\begin{subfigure}{0.5\textwidth}
  \centering
  \includegraphics[width=\textwidth]{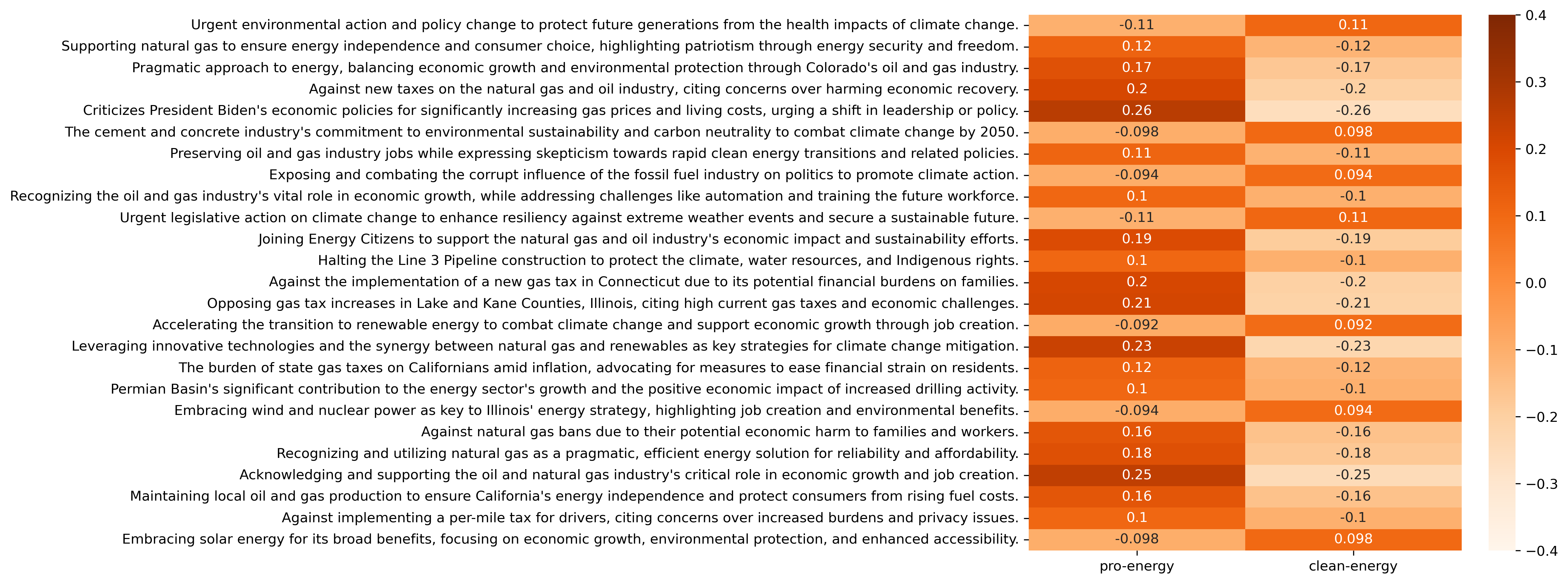}
  \caption{\textbf{Ours}: After $2^{nd}$ round of iteration (climate).}\label{fig:ours_climate}
\end{subfigure}
\caption{Correlations between arguments \& moral foundations for COVID-19; arguments \& stances for climate.}
\label{fig:correl}

\end{figure*}
\subsection{Argumentative Cohesion Comparison}

As there are previously annotated moral foundations for COVID-19 vaccine campaigns \citep{islam2022understanding} and for climate campaigns, there are annotated energy industry and climate change-related stances \citep{islam2024uncoveringthm,islam2023analysis} available. We use it for baseline analysis. We can characterize arguments/talking points as the reasons people cite to support or oppose contentious topics, such as the climate change debate or the vaccine debate. We consider assignments to be better if they are more cohesive; for example, when arguments
are more strongly correlated with specific stances or moral foundations. 

For the COVID-19 vaccine debate case study, we calculate the Pearson correlation matrices to evaluate the correlation between arguments and moral foundations, presenting the results in heatmaps (Fig. \ref{fig:correl}). 
In Fig. \ref{fig:ours_covid}, we show random $15$ arguments (among $78$ total arguments of COVID-19) to show the correlation between the arguments and the moral foundations. To compare with LDA baseline, we choose $10$ topics (Fig. \ref{fig:lda_1_covid}) and $20$ topics (Fig. \ref{fig:lda_2_covid}) to keep our $15$ arguments as middle ground. 
We can interpret reasons as distributions over moral foundations. Our analysis reveals that our arguments (Fig. \ref{fig:ours_covid}) exhibit higher correlations with moral foundations than those topics derived from LDA (Fig. \ref{fig:lda_1_covid}, Fig. \ref{fig:lda_2_covid}).
For example, in Fig. \ref{fig:ours_covid}, we see that \textbf{liberty/oppression} moral foundation is strongly correlated with the talking point “Against COVID vaccine mandates and passports, urging support for personal choice and constitutional rights regarding vaccination.” Additionally, the \textbf{purity/degradation} moral foundation is notably correlated with the talking points “Vaccination as an act of love and protection towards neighbors, aligning with Christian values.” and “Using religion as a means to exempt individuals from mandatory COVID-19 vaccination requirements.” Other expected trends emerge, such as \textbf{fairness/cheating} being highly correlated with talking points related to “Global vaccine equity, emphasizing free and universal access to COVID-19 vaccines to end the pandemic for all”.

For the climate debate case study, we perform a correlation test between the identified talking points and the stance expressed in the ads (i.e., pro-energy or clean-energy). We calculate the
Pearson correlation matrices and present them in Fig. \ref{fig:correl}. We compare our discovered talking points with a set of topics extracted using LDA. 
In Fig. \ref{fig:ours_climate}, we show random $25$ arguments (among $213$ total arguments of climate) to show the correlation between the arguments and the stance. To compare with LDA baseline, we choose $10$ topics (Fig. \ref{fig:lda_1_climate}) and $30$ topics (Fig. \ref{fig:lda_3_climate}) to keep our $25$ arguments as middle ground.
We observe that our talking points (Fig. \ref{fig:ours_climate})
have stronger correlations with stances than the derived LDA topics (Fig. \ref{fig:lda_1_climate}, Fig. \ref{fig:lda_3_climate}). For example, we note that \textbf{pro-energy} stance is strongly correlated with talking point such as “Criticizes President Biden's economic policies for significantly increasing gas prices and living costs, urging a shift in leadership or policy.”
Hyperparameters are detailed in App. \ref{hyper}.
\subsection{Ablation Study}
 We provide an ablation study in which we mine arguments from the top k instances of a cluster without summarizing them. We find comparable results (Table \ref{abla}) in terms of coverage based on the distance threshold. We use a threshold determined by the distance from the cluster centroid to an ad, with shorter distances being preferable.
 \begin{table}[h!]
\centering
\resizebox{\columnwidth}{!}{%
\begin{tabular}{l|l|cccc}
\toprule
{\bf Case}  &  {\bf \multirow{2}{*}{Iter.}}  &  \multicolumn{4}{c}{\textbf{Number of covered ads}} \\
  {\bf Study}&   & {\bf thr < 0.6} & {\bf thr < 0.5} & {\bf thr < 0.4} & {\bf thr < 0.3} \\
\midrule
\multirow{4}{*}{Climate} & 1    & 13319 & 10677 & 5355 & 1132 \\

 & 1-w/o sum.  & \multirow{1}{*}{13394} & \multirow{1}{*}{10613} & \multirow{1}{*}{5189} & \multirow{1}{*}{1164} \\

& 2    & 13669 & 11541 & \textbf{6360} & \textbf{1458}  \\

& 2-w/o sum. & \multirow{1}{*}{\textbf{13759}} & \multirow{1}{*}{\textbf{11592}} & \multirow{1}{*}{6143} & \multirow{1}{*}{1384}  \\
\midrule
\midrule
\multirow{4}{*}{COVID-19} & 1   & 7962 & 6525 & 3589 & 850  \\

& 1-w/o sum.  & 8133  & 6507 & 3477 & 787  \\

& 2   & 8197  & \textbf{6833} & \textbf{4015} & \textbf{1089}  \\

& 2-w/o sum.   & \textbf{8426}  & 6767 & 3710 & 908  \\

\bottomrule
\end{tabular}}
\caption{Ablation study (coverage). sum: summary, thr: threshold. }
\label{abla}
\end{table}
 \begin{table}
    \centering
    \begin{tabular}{lll}
    \toprule
    \textbf{\textsc{Model}} & \textbf{\textsc{Acc}} & \textbf{\textsc{F1}}  \\
    \midrule
        $Longformer_{text}$ & 90.13\% & 89.89\%\\
        $Longformer_{tp}$ & 83.43\% & 83.44\% \\
        $\mathbf{Longformer_{text+tp}}$ & \textbf{93.30\%} & \textbf{ 93.29\%} \\
        $RoBERTa_{text}$ & 93.07\% & 92.96\%\\
        $RoBERTa_{tp}$ & 82.73\% & 82.81\% \\
        $\mathbf{RoBERTa_{text+tp}}$ & \textbf{93.65\%} & \textbf{93.56\%} \\
        $llama3_{text}$ & 92.00\% & 90.95\%\\
        $llama3_{tp}$ & 81.00\% & 78.68\% \\
        $\mathbf{llama3_{text+tp}}$ & \textbf{92.50\%} & \textbf{91.49\%} \\
    \bottomrule
    \end{tabular}
    \caption{Contribution of talking point (tp) in stance classifier for climate campaigns dataset.}
    \label{tab:stance_climate}
\end{table}
\section{Downstream Task}
We design a downstream task of stance classification using the uncovered talking points. The task is
intuitively designed to learn the
association between the stances and the talking points used by the advertisers for contentious debate. 

Stance Prediction is designed as a binary classification task. In this task, the model is given a talking point uncovered by our LLMs-in-the-Loop approach and an ad text. The task is to predict the stance of the ad, i.e., \textit{pro-energy} or \textit{clean-energy}. We use pre-trained Longformer \citep{beltagy2020longformer} model and RoBERTa \citep{liu2019roberta} model for the stance classifiers. Besides, we add LLM-based classifier LLaMA \citep{touvron2023llama} for comparison and here we use llama3-70b-8192\footnote{\url{https://github.com/meta-llama/llama3}}. We measure the impact of talking points and present results in Table \ref{tab:stance_climate}.  
Our quantitative analysis in Table \ref{tab:stance_climate} is designed to demonstrate the contribution of the extracted arguments for the stance identification downstream task.
In our ablation study, we compare the performance of classifiers using only text or talking points. We find that the stance prediction task improves when talking points are included with ad text. For example: when using Longformer, macro average F1 score improves from $89.89\%$ (text only) to $93.29\%$; when using RoBERTa, it improves from $92.96\%$ (text only) to $93.56\%$; and when using llama3, it improves from $90.95\%$ (text only) to $91.49\%$ (Table \ref{tab:stance_climate}). Hyperparameters are provide in App. \ref{hyper}.
%
\section{Analysis}
This section provides a detailed analysis of how specific campaigns target demographics and adjust messaging based on real event occurrences.
\subsection{Demographic Targeting}
Understanding how messages are tailored to target particular demographic groups is crucial for opinion analysis. To analyze the variation in talking points on climate campaigns across different age groups, we examine targeted ads within three age categories, i.e., \textbf{young people} (ages $13$-$24$), \textbf{working-age people} (ages $25$-$54$), and the \textbf{older population} (age $55+$), focusing on California (CA) and Texas (TX). We choose these states because TX is recognized as the world's energy capital\footnote{\href{https://www.eia.gov/todayinenergy/detail.php?id=49356}{www.eia.gov//todayinenergy/detail.php?id=49356}}, while CA has recently positioned itself as a leading advocate in the battle against climate change\footnote{\href{https://www.pewtrusts.org/en/research-and-analysis/blogs/stateline/2022/10/06/california-takes-leading-edge-on-climate-laws-others-could-follow}{www.pewtrusts.org//en/research-and-analysis/blogs/stateline/2022/10/06/california-takes-leading-edge-on-climate-laws-others-could-follow}}. To identify the top-k most frequently mentioned entities in the texts, we prompt LLMs in a zero-shot setting. We show the results in App. \ref{demo} (Table \ref{age_climate}). 
We observe that talking points vary by age group in TX. For young people, the focus is on \textit{energy reliability and affordability, learning from Texas’ deregulation disaster}. For the working-age population, the emphasis is on \textit{accelerating the transition to renewable energy to combat climate change and support economic growth through job creation}. Meanwhile, for the older population, the discussion centers on the need for \textit{urgent legislative action on climate change}.

\begin{figure*}[ht]
\centering
\begin{subfigure}{1\textwidth}
  \centering
  \includegraphics[width=\textwidth]{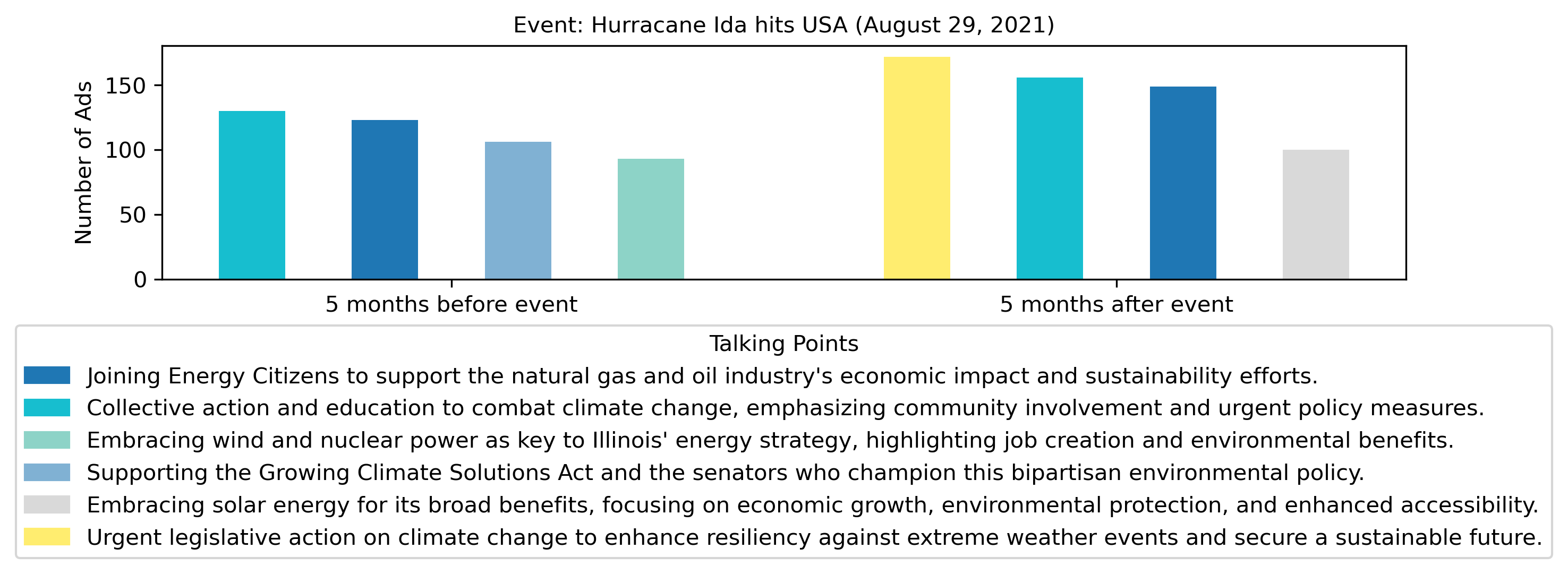}
  \caption{\textbf{Event:} Hurricane Ida hits on $29$ August $2021$.}
  \label{fig:ida}
\end{subfigure}
\begin{subfigure}{1\textwidth}
  \centering
  \includegraphics[width=\textwidth]{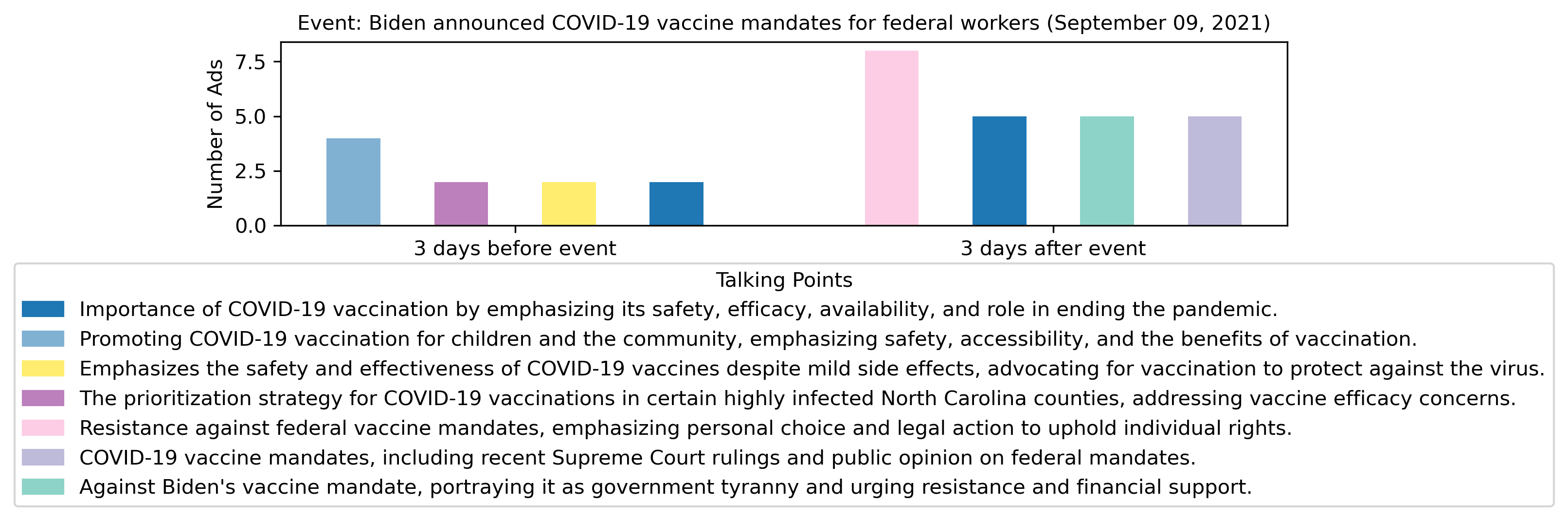}
  \caption{\textbf{Event:} Biden announced COVID-19 vaccine mandates for federal workers on $09$ September $2021$.}
  \label{fig:vax}
\end{subfigure}
\caption{The top-$4$ talking points used by advertisers before and after the events. }

\label{event}
\end{figure*}
We adopt a similar approach to analyze talking points across three different age groups in COVID-19 vaccine campaigns. We select Florida (FL) because the coronavirus surged to record levels there in August $2021$\footnote{\url{https://en.wikipedia.org/wiki/COVID-19_pandemic_in_Florida}}, and Texas (TX) because, as of April $3^{rd}$, $2021$, its vaccination rates were lagging behind the US average\footnote{\url{https://en.wikipedia.org/wiki/COVID-19_pandemic_in_Texas}}. Results are detailed in Table \ref{age_covid}. 
%
For the working age group in FL, the ad's talking point advocates for \textit{building and restoring public trust in the COVID-19 vaccine and medical community} by criticizing FL Governor Ron DeSantis's appointment of Dr. Joseph Ladapo as Surgeon General, characterizing Ladapo as an anti-masker and vaccine skeptic. The critique suggests that this appointment is insensitive or disrespectful and emphasizes the need to appoint someone who believes in science, not conspiracy theories. While targeting an older population of FL, the talking point evolves around \textit{equitable vaccine distribution and access for senior citizens} (Table \ref{age_covid}). On the other hand, while targeting a young population of TX, the talking point focuses on \textit{COVID-19 vaccine safety for children} (Table \ref{age_covid}).
\begin{table}
\centering
\small
\begin{tabular}{p{.7cm}|p{.33cm}|p{2.2cm}|p{3cm}}
\hline
 Age & \multirow{1}{*}{St.} & \multirow{1}{*}{Entity} & \multirow{1}{*}{Talking Points} \\
\hline
 \multirow{6}{*}{13-24} & \multirow{6}{*}{TX} & Children, Parents, reproductive health. & Advocates for the safety of COVID-19 vaccines for children, emphasizing mild side effects and community protection through vaccination.\\
  \hline
 \multirow{5}{*}{25-54} & \multirow{5}{*}{FL} & Ron DeSantis, Dr. Joseph Ladapo, Surgeon General. & Advocates for building and restoring public trust in the COVID-19 vaccine and the medical community.   \\
  \hline
 \multirow{14}{*}{55+} & \multirow{7}{*}{TX} & seniors, Pfizer,  who have passed away or are hospitalized due to Covid, I. & Strongly advocates for COVID-19 vaccination, highlighting its safety, efficacy, and crucial role in preventing severe illness and ending the pandemic. \\
 \cmidrule(r){2-4}
  & \multirow{5}{*}{FL} & Governor Ron DeSantis, seniors, loved one, Johnson \& Johnson. &  Efforts and challenges in equitable vaccine distribution and access for seniors across various counties.  \\
\hline
\end{tabular}
\caption{Most mentioned entities and talking points of targeted ads for \textbf{young}, \textbf{working-age}, \textbf{older} population of TX and FL from \textbf{COVID-19 vaccine campaigns} dataset. St : State.}

\label{age_covid}
\end{table}

\subsection {Shifts in Talking Points Following Significant Events}

To investigate how talking points change in response to real-world events, we pick $2$ defining events ($1$ event per case study) from our case studies. The events are as follows:
\newline
 \textbf{Hurricane Ida:} On August $29$, $2021$, Hurricane Ida made landfall near Port Fourchon, Louisiana, as a Category 4 hurricane\footnote{\url{https://en.wikipedia.org/wiki/Hurricane_Ida}}.
\newline
\textbf{Federal Vaccine Mandate:} On September $09$, $2021$, President Biden announced an executive order on COVID-19 vaccine mandates for federal workers, large employers, and health care staff\footnote{\href{https://www.whitehouse.gov/briefing-room/presidential-actions/2021/09/09/executive-order-on-requiring-coronavirus-disease-2019-vaccination-for-federal-employees/}{www.whitehouse.gov/briefing-room/presidential-actions/2021/09/09/executive-order-on-requiring-coronavirus-vaccination-for-federal-employees/}}.
Figure \ref{fig:ida} displays the top-four shifts in talking points in sponsored advertisements over the five months before and after \textbf{Hurricane Ida} struck the USA. There is a noticeable presence of arguments for "urgent legislative action on climate change to enhance resiliency against extreme
weather events and secure a sustainable future." as well as advocacy for "embracing solar energy for its broad benefits, focusing on economic growth, environmental protection, and enhanced accessibility." following the hurricane (see Figure \ref{fig:ida}).
Figure \ref{fig:vax} illustrates the talking points used in sponsored ads three days before and after \textbf{President Biden announced the federal vaccine mandate}. In the aftermath, three new predominant arguments emerged: "Resistance against federal vaccine mandates, emphasizing personal choice and legal action to uphold individual rights.", "COVID-19 vaccine mandates, including recent Supreme Court rulings and public opinion on federal mandates.", "Against Biden's vaccine mandate, portraying it as government tyranny and urging resistance and financial support." (Figure \ref{fig:vax}).
These shifts suggest that sponsored content is highly responsive to real-world events, quickly reflecting and potentially shaping public discourse on pivotal issues.
\vspace{-1pt}
\section{Conclusion}
\vspace{-2pt}
We introduce a simple yet effective iterative \textbf{LLMs-in-the-Loop} framework for uncovering latent arguments from social media messaging. Our quantitative results show that newly discovered talking points can cover a larger portion of messaging and map instances to associated arguments accurately with respect to human judgment. Besides, our stance prediction downstream task demonstrates that including talking point information helps to improve the performance of the classifier. Additionally, our analysis highlights how these talking points are specifically tailored for demographic targeting and dynamically shift in response to real-world events.
\section{Limitations}
%
Though we show the effectiveness of our framework focusing on two case studies, i.e., climate and COVID-19 vaccine campaigns, it is a domain-agnostic framework. However, the main idea of uncovering talking points using LLMs and associating the ad to their closest talking point is applicable to other cases with no changes.

Our approach relies on GPT-4 to generate a multi-document summary and talking point. We chose GPT-4 instead of the
open-source counterparts due to computational resource constraints. 

LLMs are trained on vast amounts of human-generated text, potentially embedding numerous human biases \citep{blodgett2020language,brown2020language}. We did not consider any bias in our task.

Some minority viewpoints may not form a cluster or appear in the top-k instances of a cluster to be summarized. In such cases, their viewpoints could be overlooked. In the future, we plan to refine the clusters by integrating techniques like outlier detection \citep{duan2009cluster} and feedback mechanisms to better capture and represent minority perspectives in our summaries.

A limitation of this study is the inherent subjectivity in interpreting demographic targeting, as the differentiation of themes based on demographics (specifically age and gender in this paper) is subjective, which can affect the utility of research findings.
\section{Ethics Statement}
To our knowledge, our research work, as described in this paper, adhered to all ethical guidelines. 
Instead of introducing a new dataset, we conducted experiments using two existing datasets, which have been properly cited. 
In this work, we iterated the LLMs-in-the-Loop approach twice. In the future, we aim to determine the optimal number of iterations and identify when to decide to stop exploring unassigned instances. The author's personal views are not represented in any qualitative result that we report. 
\section{Acknowledgments}
We thank the anonymous reviewers for their insightful comments. This work was partially funded by Purdue Graduate School Summer Research Grant (to TI) and NSF CAREER award IIS-2048001.
\bibliography{custom}
\appendix
\section{Appendix}
\subsection{Themes}
\label{thm}
Table \ref{tab1} shows the derived themes from the climate campaign \citep{islam2024uncoveringthm,islam2023analysis} and COVID-19 vaccine campaign \citep{islam2022understanding} datasets.
\begin{table*}
\begin{center}
 \scalebox{1}{\begin{tabular}{>{\arraybackslash}m{2cm}|>{\arraybackslash}m{13.1cm}}
 \toprule
 \textsc{\textbf{Climate}} & Economy\_pro, ClimateSolution, Pragmatism, Patriotism, AgainstClimatePolicy, 
 Economy\_clean, FutureGeneration, Environmental, HumanHealth, Animals, SupportClimatePolicy, AltEnergy, PoliticalAffiliation,
BidenGasPriceIncrease, AgainstCorporateInterests, GasTax, Deforestation, Carbon, CustomerBasedAltEnergy, EnergyAffordabilityandSustainabilityLegislation,  EcofriendlyConsumerChoices, PlasticWasteandEnvironmentalImpact,  PromoteSustainableTransportation, WaterManagementandSustainability, FoodSecurity.\\
 \hline
 \hline
 \textsc{\textbf{COVID-19}} & GovDistrust, GovTrust, VaccineRollout, VaccineSymptom, VaccineEquity, VaccineStatus, EncourageVaccination, VaccineMandate, VaccineReligion, VaccineEfficacy, VaccineDevelopment, CovidPlan, VaccineMisinformation, NaturalImmunity.\\
 \bottomrule
\end{tabular}}
\end{center}
\caption{Derived themes from previous studies: Climate campaigns \citep{islam2024uncoveringthm,islam2023analysis}, COVID-19 vaccine campaigns \citep{islam2022understanding}}
\label{tab1}
\end{table*}
\subsection{Data Statistics}
\label{data_stat}
\begin{figure*}[htp]

\sbox\twosubbox{%
  \resizebox{\dimexpr.9\textwidth-1em}{!}{%
    \includegraphics[height=3cm]{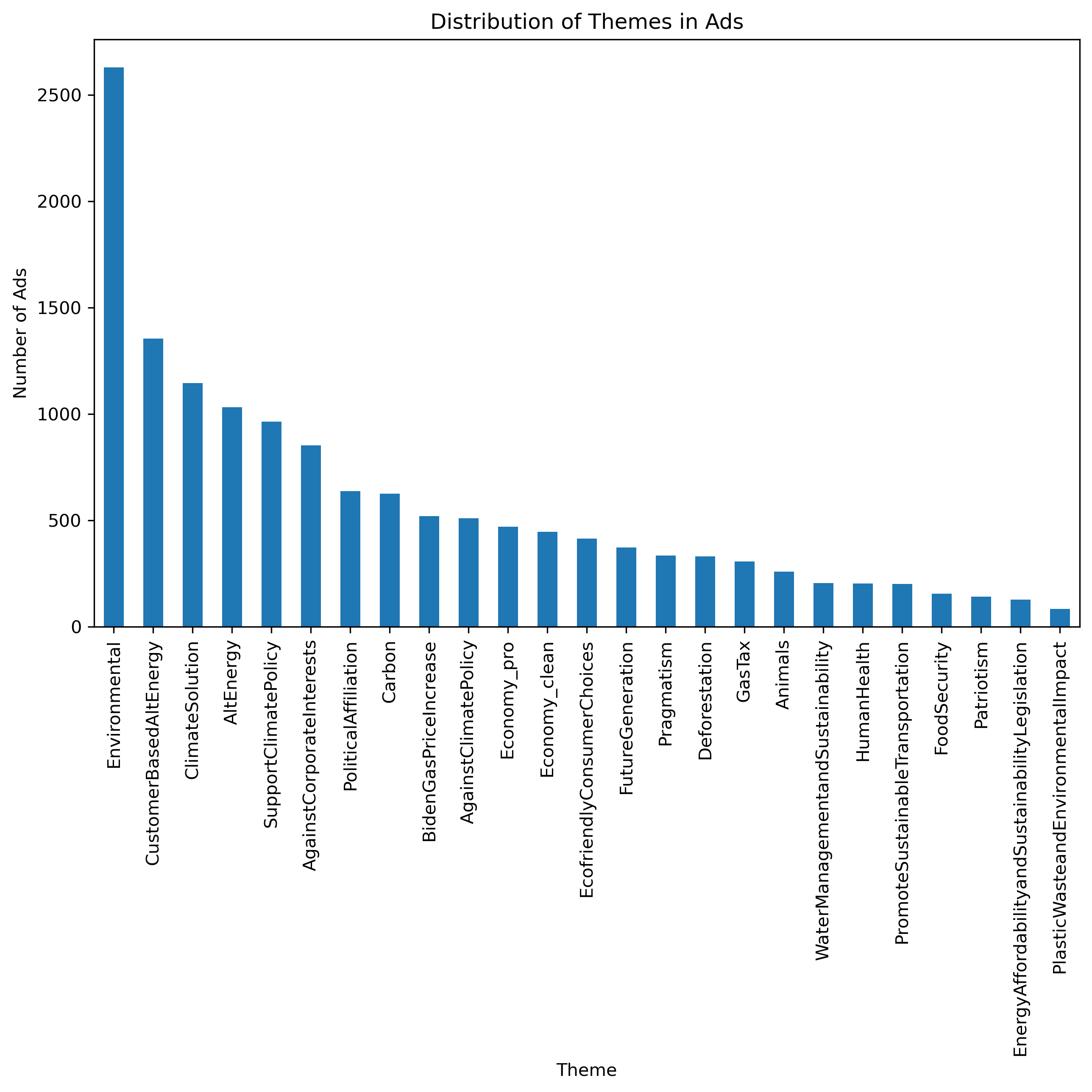}%
    \includegraphics[height=3cm]{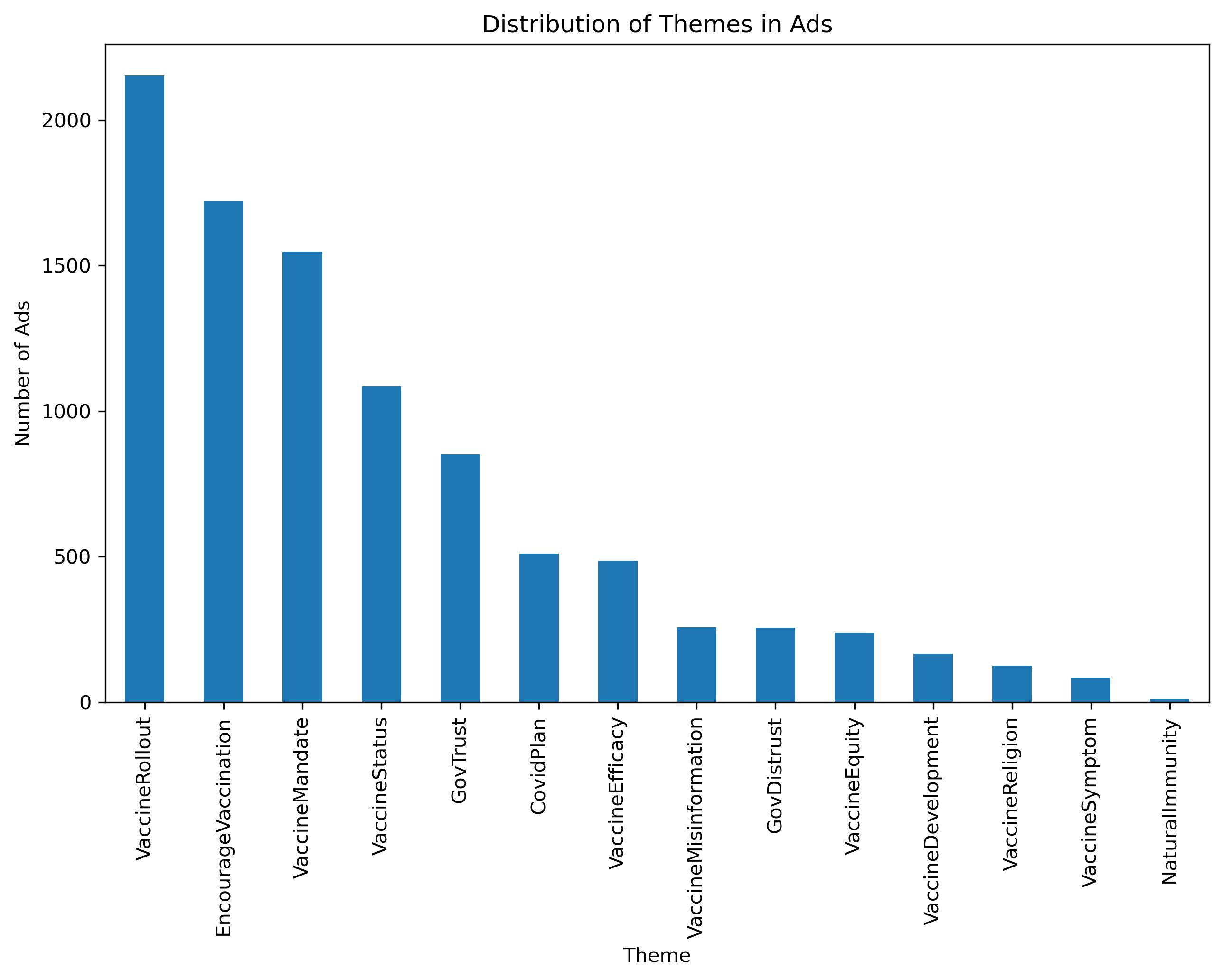}%
  }%
}
\setlength{\twosubht}{\ht\twosubbox}


\centering

\subcaptionbox{Climate campaign.\label{stat_climate}}{%
  \includegraphics[height=\twosubht]{thm_ads_dis_climate_new.png}%
}\quad
\subcaptionbox{COVID-19 vaccine campaign.\label{stat_covid}}{%
  \includegraphics[height=\twosubht]{thm_ads_dis_covid_new.png}%
}
\caption{Number of instances in each theme for climate and COVID-19 vaccine campaign.}
\label{stat}
\end{figure*}

Fig. \ref{stat} shows the statistics regarding the number of instances in each theme for climate (Fig. \ref{stat_climate}) and COVID-19 vaccine (Fig. \ref{stat_covid}) campaign. In Table \ref{sub_stat_climate}, we show the number of sub-clusters identified under each theme for both iterations of the climate case. In Table \ref{sub_stat_covid}, we provide the number of sub-clusters identified under each theme for both iterations of the COVID-19 case study.
\begin{table*}
\centering
\resizebox{\textwidth}{!}{%
\begin{tabular}{|l|c|c|}
\hline
\textbf{Climate theme} & \textbf{\# sub-clusters per theme (Iter1)} & \textbf{\# sub-clusters per theme (Iter2)} \\ \hline
Economy\_pro & 6 & 4 \\ \hline
ClimateSolution & 3 & 2 \\ \hline
Pragmatism & 8 & 2 \\ \hline
Patriotism & 4 & 3 \\ \hline
AgainstClimatePolicy & 3 & 2 \\ \hline
Economy\_clean & 4 & 6 \\ \hline
HumanHealth & 3 & 7 \\ \hline
FutureGeneration & 4 & 4 \\ \hline
Environmental & 3 & 6 \\ \hline
Animals & 5 & 5 \\ \hline
AltEnergy & 5 & 7 \\ \hline
SupportClimatePolicy & 7 & 4 \\ \hline
PoliticalAffiliation & 5 & 4 \\ \hline
BidenGasPriceIncrease & 3 & 3 \\ \hline
AgainstCorporateInterests & 7 & 3 \\ \hline
GasTax & 7 & 3 \\ \hline
Deforestation & 7 & 6 \\ \hline
Carbon & 6 & 3 \\ \hline
CustomerBasedAltEnergy & 7 & 5 \\ \hline
FoodSecurity & 4 & 5 \\ \hline
EnergyAffordabilityandSustainabilityLegislation & 4 & 5 \\ \hline
EcofriendlyConsumerChoices & 5 & 3 \\ \hline
PlasticWasteandEnvironmentalImpact & 6 & 1 \\ \hline
PromoteSustainableTransportation & 6 & 4 \\ \hline
WaterManagementandSustainability & 4 & 4 \\ \hline
\end{tabular}}
\caption{Number of sub-clusters per climate theme across two iterations.}
\label{sub_stat_climate}
\end{table*}
\begin{table*}
\centering
\resizebox{\textwidth}{!}{%
\begin{tabular}{|l|c|c|}
\hline
\textbf{COVID-19 theme} & \textbf{\# sub-clusters per theme (Iter1)} & \textbf{\# sub-clusters per theme (Iter2)} \\ \hline
VaccineRollout & 5 & 2 \\ \hline
EncourageVaccination & 3 & 2 \\ \hline
VaccineMandate & 6 & 4 \\ \hline
VaccineStatus & 4 & 1 \\ \hline
GovTrust & 2 & 3 \\ \hline
CovidPlan & 3 & 2 \\ \hline
VaccineEfficacy & 4 & 2 \\ \hline
VaccineMisinformation & 2 & 4 \\ \hline
GovDistrust & 3 & 2 \\ \hline
VaccineEquity & 5 & 2 \\ \hline
VaccineDevelopment & 3 & N/A \\ \hline
VaccineReligion & 2 & 3 \\ \hline
VaccineSymptom & 4 & 4 \\ \hline
NaturalImmunity & 2 & 1 \\ \hline
\end{tabular}}
\caption{Number of sub-clusters per COVID-19 theme across two iterations.}
\label{sub_stat_covid}
\end{table*}

\begin{table*}[htb]
\begin{center}
\begin{tabular}{l|c|l}
\toprule
{\bf Case }  & \multirow{2}{*}{\bf Iter.}  & \multirow{2}{*}{\bf Arguments} \\
{\bf Study} &  & \\
\midrule
\multirow{18}{*}{\textbf{Climate}} & \multirow{10}{*}{1}   & (1) Prioritizing domestic energy production as a patriotic duty to \\ & ~ & ensure America's independence, security, and economic prosperity. \\
     & ~ & (2) Maintaining local oil and gas production to ensure economic and \\ & ~ & energy security, highlighting patriotism through self-reliance.\\
     & ~ & (3) Supporting the oil and gas industry as a patriotic duty,\\ & ~ & emphasizing its critical role in national security and economic health. \\
  &  ~ & (4) Supporting natural gas to ensure energy independence and \\ & ~ & consumer choice, highlighting patriotism through energy security \\ & ~ & and freedom.  \\
   \cmidrule{2-3}
& \multirow{8}{*}{2}   &  (5) Rallying around patriotism, energy independence, and \\ & ~ & conservative values to protect and promote American interests \\ & ~ & and security.\\
   &  ~ & (6) Maintaining local oil and gas production to ensure California's \\ & ~ & energy independence and protect consumers from rising fuel costs.\\
   & ~ &  (7) Energy independence for Israel through faith-inspired and \\ & ~ & scientifically supported oil and gas exploration and diverse \\ & ~ & leadership in community engagement.   \\
   \midrule
    \midrule
    \multirow{16}{*}{\textbf{COVID}} & \multirow{11}{*}{1}   & (1) Against price control legislation, emphasizing its potential to \\ \multirow{16}{*}{\textbf{-19}} & ~ &  hinder vaccine development and equitable access to life-saving \\  & ~ & COVID-19 vaccinations. \\
     & ~ & (2) Global vaccine equity, emphasizing free and universal access  \\  & ~ & to COVID-19 vaccines to end the pandemic for all.\\
     & ~ & (3) Overcoming barriers to ensure equitable COVID-19 vaccine \\  & ~ & access for underserved communities and prioritizing vulnerable \\  & ~ &  populations. \\
  &  ~ & (4) Equitable access to COVID-19 vaccines, addressing eligibility, \\  & ~ & distribution fairness.  \\
   \cmidrule{2-3}
& \multirow{4}{*}{2}   &  (5) Donating funds that will be matched to multiply impact, \\  & ~ & promoting fair global access to vaccines.\\
   &  ~ & (6) Efforts and challenges in equitable vaccine distribution and \\  & ~ & access for seniors across various counties.\\
\bottomrule
\end{tabular}
\end{center}
\caption{Example of resulting arguments after $1^{st}$ and $2^{nd}$ iteration. For climate campaigns, arguments are shown under the `Patriotism' theme. For COVID-19 vaccine campaigns, arguments are shown under the `VaccineEquity' theme. All themes derived by previous studies are shown in App. \ref{thm} (Table \ref{tab1})}
\label{args}
\end{table*}
\subsection{Experimental Details}
\label{hyper}
To obtain topics from LDA, we use Gensim \cite{rehurek2011gensim} implementation. We follow
the pre-processing steps shown in \citet{hoyle2021automated} and estimate the number of topics in a data-driven manner by maximizing. We do a grid search over a set of \{10, 20, 25, 30\} for the LDA baseline.

For SBERT embedding, we use the sentence transformer model `all-mpnet-base-v2' with default parameters.

For the stance classifier, we randomly split the data into three subsets, namely training set ($60\%$), validation set ($20\%$), and
test set ($20\%$). Show the result on the test set in Table \ref{tab:stance_climate}. For Longformer, we use `longformer-base-4096' model, batch size = $8$, epoch = $3$, learning rate = $2e-5$, weight decay = $0.01$. For RoBERTa, we use `roberta-base' model with same batch size, epoch, learning rate, and weight decay used for Longformer.
Our early stopping criterion is the lowest validation loss. To run the stance classifier, we use a single GPU GeForce GTX $1080$ Ti
GPU, with $4$ Intel Core $i5$-$7400$ CPU @ $3.00$ GHz processors, and it takes around $10$ minutes to run the model. 

To run llama3-70b-8192, we use Groq API\footnote{\url{https://wow.groq.com/}}. For tokenizer, we use Meta-Llama-3-70B-Instruct from Hugging Face.
\begin{table*}[t]
\centering
\begin{tabular}{p{1cm}|c|p{3.5cm}|p{6.9cm}}
\hline
 Age Group & \multirow{2}{*}{State} & \multirow{2}{*}{Entity} & \multirow{2}{*}{Talking Points} \\
\hline
 \multirow{12}{*}{13-24} & \multirow{6}{*}{CA} & Young People, Climate Change Denier, Climate Movement. & (1) Climate change's health impacts through informed initiatives and equity-focused efforts to protect vulnerable communities. (2) Collective action and education to combat climate change, emphasizing community involvement and urgent policy measures.\\
 \cmidrule(r){2-4}
  & \multirow{5}{*}{TX} & Natural Gas Systems, Coal Energy Systems, Nuclear Energy Systems, Wind Turbines, Solar Panels. & (1) Ensure energy reliability and affordability, learning from Texas' deregulation disaster.  \\
  \hline
  \hline
 \multirow{9}{*}{25-54} & \multirow{5}{*}{CA} & Carbon Emissions, Sacramento Municipal Utility District. & (1) Public engagement to support and expand carbon-free energy as a crucial step against climate change. (2) Transitioning from coal to cleaner energy sources to combat climate change and promote sustainability.  \\
 \cmidrule(r){2-4}
  & \multirow{4}{*}{TX} & Climate Change, Renewable Energy Plan. & (1) Accelerating the transition to renewable energy to combat climate change and support economic growth through job creation.  \\
  \hline
  \hline
 \multirow{7}{*}{55+} & \multirow{4}{*}{CA} & Western Burrowing Owls, Aramis Industrial Solar Power Plant. & (1) Continued investment in climate and clean energy solutions to protect birds from climate change impacts.\\
 \cmidrule(r){2-4}
  & \multirow{3}{*}{TX} & U.S. Senators or Congressman, Citizens' Climate Lobby. & (1) Urgent legislative action on climate change to enhance resiliency against extreme weather events and secure a sustainable future. \\
\hline
\end{tabular}
\caption{Most mentioned entities and talking points of targeted ads for \textbf{young}, \textbf{working-age}, \textbf{older} population of CA and TX from \textbf{climate campaigns} dataset.}
\label{age_climate}
\end{table*}
\subsection{Arguments under specific Themes}
\label{arg_thm}
Table \ref{args} shows examples of resulting arguments after $1^{st}$ and $2^{nd}$ iteration under the `Patriotism' theme for climate campaigns and the `VaccineEquity' theme for COVID-19 vaccine campaigns theme.

\subsection{Resulting Arguments}
\label{arg_final}
In this section, we provide the resulting arguments using our LLMs-in-the-Loop approach for both datasets.
\subsubsection{Climate Campaigns}
Table \ref{arg_climate} provides the resulting arguments related to climate campaigns after two rounds of iterations of our LLMs-in-the-Loop approach.
\subsubsection{COVID-19 Vaccine Campaigns}
Table \ref{arg_covid} provides the resulting arguments related to COVID-19 vaccine campaigns after two rounds of iterations of our LLMs-in-the-Loop approach.
\subsection{Human Validation}
\label{human}
Three NLP and CSS researchers, age range $30$-$45$, $2$ males and $1$ female, did two $1$-hour sessions to validate whether the generated and refined talking points as well as to annotate whether $ads \rightarrow arguments $ mapping is correct or not. The annotators included advanced graduate students and faculty.
\subsection{Demographic Analysis}
\label{demo}
Table \ref{age_climate} shows the most mentioned entities and talking points of targeted ads for \textbf{young}, \textbf{working-age}, \textbf{older} population of CA and TX from \textbf{climate campaigns} dataset. 
\onecolumn
\begin{longtable}{p{3.5cm}|p{9.7cm}}
\caption{Resulting arguments related to Climate campaigns after two rounds of iterations.} \label{arg_climate} \\
\hline
Themes & Talking Points \\
\hline
\endfirsthead
\multicolumn{2}{c}%
{{\bfseries Table \thetable\ continued from previous page}} \\
\hline
Themes & Talking Points \\
\hline
\endhead
\hline \multicolumn{2}{r}{{Continued on next page}} \\ 
\endfoot

\hline
\endlastfoot
Economy\_pro & "Against natural gas bans due to their potential economic harm to families and workers.",
    "Joining Energy Citizens to support the natural gas and oil industry's economic impact and sustainability efforts.",
    "Acknowledging and supporting the oil and natural gas industry's critical role in economic growth and job creation.",
    "Permian Basin's significant contribution to the energy sector's growth and the positive economic impact of increased drilling activity.",
    "Considering the economic and community impact of taxation on the energy sector, emphasizing support for energy-related jobs and services.",
    "The Line 3 Replacement project's role in boosting Minnesota's economy through job creation, supporting Tribal communities, and enhancing safety, while positively impacting the environment.",
    "Recognizing the oil and gas industry's vital role in economic growth, while addressing challenges like automation and training the future workforce." \\
\hline
ClimateSolution & "Adopting propane as a versatile and environmentally friendly energy solution to support climate change mitigation efforts.",
    "Leveraging innovative technologies and the synergy between natural gas and renewables as key strategies for climate change mitigation.",
    "Leveraging diverse, innovative, and market-based strategies to effectively address climate change and promote environmental sustainability.",
    "Adoption of RGGI in Pennsylvania, international cooperation on climate action, and exploring market-based, conservative solutions for sustainable economic and environmental health.",
    "Investing in innovative technologies like ICHOR's RKI Injector and AT\&T's initiatives to drastically reduce emissions and improve fuel efficiency." \\
    \hline
    Pragmatism & "Pragmatic, multi-tiered action on climate change, emphasizing renewable energy investments and policy for sustainable development and job growth.",
    "Practical investments in sustainable infrastructure and energy diversity to address climate change and ensure future resilience.",
    "Pragmatic approach to energy transition, emphasizing the need to harmonize environmental sustainability, economic growth, and job preservation.",
    "Recognizing and utilizing natural gas as a pragmatic, efficient energy solution for reliability and affordability.",
    "Pragmatic approach to energy policy that balances clean energy transition with economic growth and job preservation.",
    "Pragmatic approach to energy, balancing economic growth and environmental protection through Colorado's oil and gas industry.",
    "Pragmatic approach to energy, emphasizing the need for oil and safe pipeline infrastructure for global energy security.",
    "Pragmatic policy changes and discussions that address workers' rights, environmental sustainability, and societal challenges for improvement and growth.",
    "Pragmatic solutions to environmental challenges through technological innovation, collaboration, and strategic policy efforts to ensure sustainability and security."\\
    \hline
Patriotism & "Prioritizing domestic energy production as a patriotic duty to ensure America's independence, security, and economic prosperity.",
     "Maintaining local oil and gas production to ensure economic and energy security, highlighting patriotism through self-reliance.",
     "Supporting the oil and gas industry as a patriotic duty, emphasizing its critical role in national security and economic health.",
     "Supporting natural gas to ensure energy independence and consumer choice, highlighting patriotism through energy security and freedom.",
     "Rallying around patriotism, energy independence, and conservative values to protect and promote American interests and security.",
     "Maintaining local oil and gas production to ensure California's energy independence and protect consumers from rising fuel costs.",
     "Energy independence for Israel through faith-inspired and scientifically supported oil and gas exploration, and diverse leadership in community engagement." \\
     \hline
     AgainstClimatePolicy & "Renewable energy investment as a strategy for economic growth and job creation while addressing climate change.",
    "Transitioning to a clean energy economy, emphasizing job creation, sustainability, and combating climate change for economic recovery.",
    "Significant investment in clean energy infrastructure to create jobs and power homes sustainably, driving economic and environmental progress.",
    "Passage of the bipartisan infrastructure deal to enhance clean water, transit, and create numerous good jobs.",
    "American Jobs Plan as a catalyst for economic revitalization through clean energy, creating jobs, and improving infrastructure.",
    "Enhancing sustainability through clean energy initiatives, education, infrastructure improvement, and worker health and safety in the clean energy sector.",
    "Local and state-level initiatives that promote clean energy, sustainability, and environmental protection to build a more sustainable future.",
    "Comprehensive community and environmental betterment through sustainable economic practices, fair wages, and innovative climate solutions.",
    "Rigorous scrutiny and equitable policies in clean-energy projects to ensure they genuinely benefit the environment and communities."\\
    Economy\_clean &  "Renewable energy investment as a strategy for economic growth and job creation while addressing climate change.",
    "Transitioning to a clean energy economy, emphasizing job creation, sustainability, and combating climate change for economic recovery.",
    "Significant investment in clean energy infrastructure to create jobs and power homes sustainably, driving economic and environmental progress.",
    "Passage of the bipartisan infrastructure deal to enhance clean water, transit, and create numerous good jobs.",
    "American Jobs Plan as a catalyst for economic revitalization through clean energy, creating jobs, and improving infrastructure.",
    "Enhancing sustainability through clean energy initiatives, education, infrastructure improvement, and worker health and safety in the clean energy sector.",
    "Local and state-level initiatives that promote clean energy, sustainability, and environmental protection to build a more sustainable future.",
    "Comprehensive community and environmental betterment through sustainable economic practices, fair wages, and innovative climate solutions.",
    "Rigorous scrutiny and equitable policies in clean-energy projects to ensure they genuinely benefit the environment and communities." \\
    \hline
    HumanHealth & "Reducing air pollution from oil and gas industries to protect children's health and ensure environmental justice.",
    "Electing knowledgeable leaders and supporting organizations to address climate change's significant health and humanitarian impacts.",
    "Climate change's health impacts through informed initiatives and equity-focused efforts to protect vulnerable communities.",
    "Leadership that understands and addresses the health impacts of climate change, emphasizing immediate action for global health and stability.",
    "Environmental health risks, from fracking and gas stoves to substance abuse, to protect and improve public health.",
    "Combining environmental activism with health equity to combat pollution and support sustainable, just communities.",
    "Awareness of gas stoves' health risks, promoting safer alternatives, and implementing preventive measures to protect indoor air quality and prevent asthma.",
    "Informed, compassionate leadership and global medical support to address healthcare challenges, ensuring dignity and comprehensive care across life stages and crises.",
    "Addressing and mitigating air pollution sources to protect public health and promote environmental sustainability.",
    "Proactive engagement and informed responses to health and safety challenges exacerbated by environmental changes and societal issues." \\
    \hline
    FutureGeneration & "Urgent environmental action and policy change to protect future generations from the health impacts of climate change.",
    "Educating children about climate change through open discussions to inspire future generations to take action for their planet's health.",
    "Collective action led by motivated parents to combat climate change for a healthier future for children.",
    "Empowering the youth to lead climate change solutions, emphasizing education, activism, and support for vulnerable communities.",
    "Civic engagement and local policy reforms to secure a safer, fairer, and sustainable future for the next generation.",
    "Sustainable and renewable energy initiatives led by Indigenous and community efforts to ensure a healthier planet for future generations.",
    "Integrated solutions to climate change that also address social, racial, and economic inequities for future generations' benefit.",
    "Sustainable energy and community initiatives that create educational, economic, and environmental benefits for current and future generations." \\
    \hline
    Environmental & "Ending fossil fuel use, implementing clean energy transitions, and enacting protective regulations to combat climate change and pollution.",
    "Urgent legislative action on climate change to enhance resiliency against extreme weather events and secure a sustainable future.",
    "Collective action and education to combat climate change, emphasizing community involvement and urgent policy measures.",
    "Community resilience through local initiatives, transitioning to clean energy, sustainable practices, and comprehensive support for businesses impacted by environmental disasters.",
    "Environmental protection and Indigenous rights by opposing oil pipelines that threaten natural resources and exacerbate the MMIW epidemic.",
    "Vigilance and community action against organized crime targeting green energy resources, and promotes sustainable living practices.",
    "Proactive climate change solutions through innovative forestry, ecosystem restoration projects, clean energy investment, and urgent policy actions to safeguard the future.",
    "Climate change's role in exacerbating droughts, fires, and heatwaves, underscored by scientific consensus and observable impacts like Minnesota's rising temperatures.",
    "Active participation in environmental conservation efforts led by Derek Kilmer to restore Puget Sound, protect wilderness, and combat climate change." \\
    \hline
    Animals & "Re-protection of gray wolves, recognizing their vital ecological role, and calls for increased conservation efforts and public support.",
    "Urgent financial contributions, leveraging donation matches to combat threats to endangered wildlife like habitat loss and climate change.",
    "Urgent climate action to protect Alaska's fisheries, crucial for both the economy and marine ecosystems.",
    "Continued investment in climate and clean energy solutions to protect birds from climate change impacts.",
    "Immediate public action to conserve wildlife and address climate change's role in accelerating species extinction.",
    "Habitat restoration and conservation to support wildlife, enhance outdoor activities, and address climate change for future ecological balance.",
    "Protecting and restoring ocean ecosystems to preserve marine biodiversity and combat climate change's impacts on animals.",
    "Awareness and action against the environmental and ethical harms of animal agriculture, emphasizing sustainable choices and animal rights.",
    "Habitat conservation, creating pollinator gardens, and innovative habitat restoration to protect bats, bees, and wildlife crucial for ecosystem health and biodiversity.",
    "Promoting renewable energy projects as a means to generate clean power, support local economies, and protect marine biodiversity for future generations." \\
    \hline
    AltEnergy & "Accelerating the transition to renewable energy to combat climate change and support economic growth through job creation.",
    "Promoting clean and renewable energy to foster economic growth, create jobs, and address climate change effectively.",
    "Public engagement to support and expand carbon-free energy as a crucial step against climate change.",
    "The adoption of wind energy as a crucial, effective solution for reducing CO2 emissions and combating climate change.",
    "Diversified renewable energy solutions to ensure sustainability, grid reliability, and meet increasing demand for clean power across different regions.",
    "Expansion and innovation in bioethanol production and solar energy to support a sustainable, clean energy future.",
    "The positive impact of wind energy, emphasizing job creation, economic growth, environmental benefits, and community engagement.",
    "Political support and action towards clean energy, job creation, climate action, and sustainability initiatives in local communities.",
    "Investment in ICHOR's revolutionary fuel injection technology to reduce emissions and fuel consumption.",
    "Transitioning from coal to cleaner energy sources to combat climate change and promote sustainability.",
    "The adoption and celebration of renewable energy initiatives while balancing environmental preservation." \\
    \hline
    SupportClimatePolicy & "Passing the Build Back Better Act, highlighting its importance for economic growth, job creation, clean energy employment, and combating climate change.",
    "Supporting the Growing Climate Solutions Act and the senators who champion this bipartisan environmental policy.",
    "Supporting the Growing Climate Solutions Act as a means to boost farmer income and combat climate change.",
    "Legislative action and public support for policies that address climate change through clean energy investment and promote social justice.",
    "The passage of the Build Back Better Act, emphasizing its investments in clean energy and climate action.",
    "Halting the Line 3 Pipeline construction to protect the climate, water resources, and Indigenous rights.",
    "Voting for candidates who prioritize climate change action and sustainable policies in their platforms.",
    "The inclusion of bold climate action in legislation, simultaneous with job creation and equity." \\
    \hline
    PoliticalAffiliation & "Active political participation and advocacy to influence policies on environmental protection and social justice across various states.",
    "Exposing and combating the corrupt influence of the fossil fuel industry on politics to promote climate action.",
    "Appointments that prioritize environmental protection and resist the fossil fuel industry's influence in government and finance.",
    "Holding fossil fuel companies accountable through political, public action, and legislative reforms for their environmental impact and misinformation.",
    "Greater scrutiny of political ties to the oil and gas industry and calls for fair regulation.",
    "Understanding and challenging perceived extremism in politics, scrutinizing shifts in political power and ideology, and recognizing ongoing political efforts on issues like infrastructure and climate change.",
    "Supporting candidates whose platforms align with individual voter values and community needs.",
    "Political appointments and decisions that prioritize climate action and reduce the influence of fossil fuels." \\
    \hline
    BidenGasPriceIncrease & "Reversing President Biden's energy policies to support American energy production and reduce gas prices for families.",
    "Against perceived financial mismanagement by 'DC liberals', linking it to inflation and higher gas prices under Biden's presidency.",
    "Mobilizing against government mandates perceived to increase gas prices, emphasizing grassroots engagement to protect consumer interests.",
    "Opposing a spending plan criticized for contributing to higher gas prices and economic burdens.",
    "Criticizes President Biden's economic policies for significantly increasing gas prices and living costs, urging a shift in leadership or policy.",
    "A return to energy independence to control gas prices and criticizes current policies for rising fuel costs." \\
    \hline
    AgainstCorpInter & "Supporting Sarah, who prioritizes climate action, honesty, and fiscal responsibility over corporate interests and misinformation.",
    "Grassroots support over corporate interests, emphasizing financial contributions to champion climate, democracy, and economic issues in Iowa's race.",
    "Grassroots-supported candidates promoting progressive policies over those funded by corporate interests and lobbyists in Congressional races.",
    "Dismantling financial and political support for the fossil fuel industry to favor sustainable, clean energy solutions and accountability.",
    "Electing progressive candidates who prioritize public welfare and grassroots support over corporate interests and profits.",
    "Using the power of voting to challenge corporate interests by supporting policies and candidates focused on public welfare and environmental sustainability.",
    "Empowering grassroots movements over corporate interests, emphasizing small donations to drive significant policy changes for social and environmental justice.",
    "Active resistance against corporate influence in politics and society to achieve environmental, social, and economic reforms.",
    "Grassroots, community-focused leadership challenging established politics and corporate dominance to achieve progressive social and environmental reforms.",
    "Grassroots funding to counteract corporate-funded campaigns, emphasizing the power of collective small donations against big money interests." \\
    \hline
    GasTax & "Reassessment of gas tax reliance, emphasizing modernizing revenue models to sustainably fund road maintenance amidst electric vehicle challenges.",
  "Halting proposed gas tax hikes to alleviate financial burdens on residents amidst concerns over rising living costs.",
  "Repealing the CPI portion of the gas tax law to alleviate economic burdens on citizens.",
  "Against implementing a per-mile tax for drivers, citing concerns over increased burdens and privacy issues.",
  "Opposing gas tax increases in Lake and Kane Counties, Illinois, citing high current gas taxes and economic challenges.",
  "Against a proposed fuel mandate in Iowa, framed as a tax on consumers leading to increased gas prices.",
  "Against the implementation of a new gas tax in Connecticut due to its potential financial burdens on families.",
  "The burden of state gas taxes on Californians amid inflation, advocating for measures to ease financial strain on residents.",
  "Lowering gas taxes as part of broader fiscal reforms to reduce the financial burden on citizens.",
  "Emphasizes the need for transparency, accountability, and community involvement in decisions impacting gas station operations and related policies." \\
  Deforestation & "Roadless protections in the Tongass National Forest to preserve its ecosystem, combat climate change, and support local economies and cultures.",
    "Leveraging consumer purchases for tree planting to combat deforestation and climate change, emphasizing individual contributions to collective environmental efforts.",
    "Global reforestation and individual action against climate change through tree planting and support for sustainable community practices to combat deforestation.",
    "Permanent legislative protection of the Arctic National Wildlife Refuge to preserve biodiversity and prevent environmental degradation from oil and gas extraction.",
    "Preserving urban green spaces and mature trees to maintain air quality, carbon sequestration, and historical landscapes against deforestation for construction projects.",
    "Combating deforestation and climate change through community-driven reforestation efforts and financial support in California.",
    "Stopping the use of forests as fuel for electricity and urging significant climate action to address and mitigate deforestation.",
    "Halting energy projects that cause deforestation, respecting treaties, and protecting ecosystems against corporate environmental damage.",
    "Eliminating conflict palm oil from products to protect rainforests and wildlife, promoting transparency and sustainable sourcing.",
    "Improved forest management and proactive measures like controlled burns to mitigate wildfire risks and prevent deforestation.",
    "Conservation, restoration efforts, and responsible management policies to combat climate change and prevent deforestation.",
    "Restoration and protection of critical habitats to combat deforestation and support ecosystem resilience and biodiversity.",
    "Indigenous-led efforts to protect natural resources and uphold sovereignty as crucial in combating deforestation and environmental degradation." \\
    \hline
    Carbon & "The cement and concrete industry's commitment to environmental sustainability and carbon neutrality to combat climate change by 2050.",
    "Accessible carbon offset solutions, emphasizing affordability, simplicity, and the importance of taking action for Earth Day.",
    "Using Ecologi's platform to offset carbon footprint, plant trees, and collectively combat climate change.",
    "Implementing carbon pricing strategies through legislative action, communication, and collective efforts to address climate change.",
    "Collaborative efforts across various sectors to reduce carbon emissions and transition towards a sustainable, low-carbon economy.",
    "Innovative technologies and solutions to capture, reduce, and offset carbon emissions for a more sustainable and carbon-neutral future.",
    "Reducing carbon footprints through lifestyle changes, sustainable eating habits, and supporting initiatives aimed at carbon neutrality and environmental sustainability." \\
    \hline
    CustBasedAltEnergy & "Customer engagement in alternative energy solutions, emphasizing renewable and nuclear options to combat climate change and promote sustainability.",
    "Public support and awareness of nuclear energy as a sustainable, carbon-free alternative energy solution in the U.S.",
    "Community engagement and support for wind energy, highlighting its benefits for jobs, rural economies, and sustainable development.",
    "Community engagement in renewable energy through a cost-effective, equipment-free Community Solar program with enrollment incentives.",
    "Homeowner participation in solar energy trials to foster renewable energy adoption and reduce individual energy costs.",
    "Embracing solar energy for its broad benefits, focusing on economic growth, environmental protection, and enhanced accessibility.",
    "The adoption and support of utility-scale solar projects to supply clean energy to the grid for consumers.",
    "Embracing wind and nuclear power as key to Illinois' energy strategy, highlighting job creation and environmental benefits.",
    "Adopting biofuels and innovative fuel technologies to reduce emissions and promote sustainability in transportation for consumers.",
    "Community engagement and legal actions to promote sustainable development and energy practices at the local level.",
    "Adopting geothermal heating and cooling as a cost-effective, efficient, and environmentally friendly alternative energy solution for homeowners." \\
    \hline
    FoodSecurity & "Addressing climate change's impact on agriculture, social equity, and women's rights to ensure food security for all.",
    "Cross-sector collaboration, addressing food loss, engaging communities, and prioritizing food security initiatives to combat global hunger.",
    "Participation in the Ration Challenge to support emergency food aid and attending a forum to explore food security issues.",
    "Engagement with the Fruitful Communities initiative to combat climate change, promote sustainable agriculture, and enhance food security.",
    "Urgent humanitarian aid and community support to address immediate and long-term food security challenges in crisis-affected areas.",
    "Sustainable dietary changes, particularly reducing meat consumption, to improve environmental health and ensure long-term food security.",
    "Community support and mobilization to provide essential resources and aid, ensuring food security and health during crises.",
    "Combining sustainable agriculture, traditional knowledge, and community efforts to improve food security and environmental sustainability." \\
    EnrgyAffrd\&SustainLeg &  "Legislative action in Virginia to ensure energy reliability and affordability, learning from Texas' deregulation disaster.",
   "Legislative action to significantly lower energy costs, promoting cleaner and more reliable energy solutions in North Carolina.",
   "Legislative and policy efforts to enhance energy efficiency, grid reliability, and sustainability across various regions.",
   "Legislative supporting utility bill assistance, pollution reduction, energy-efficient building codes, and transition to pollution-free buildings for sustainability.",
   "Policy-driven support for renewable energy, affordability, emissions reduction, and sustainable development, emphasizing community involvement and legislative backing.",
   "Legislative action to ensure utility service protection, improve energy infrastructure resilience, and address affordability amid extreme weather and economic challenges.",
   "Transitioning to sustainable energy sources, enhancing security for clean energy businesses, and supporting policies for cleaner, affordable energy solutions.",
   "Leveraging legislation and community programs to provide financial assistance for utility bills, ensuring energy affordability for all residents." \\
   \hline
   EcofrndlyConsmChoice &  "Making eco-friendly consumer choices by purchasing sustainable apparel and actively contributing to environmental conservation efforts.",
    "Choosing eco-friendly, sustainable laundry solutions like Earthbreeze to reduce environmental impact and support ethical consumption.",
    "Adopting home water treatment systems to reduce reliance on single-use plastic bottles, promoting environmental sustainability and health.",
    "Making eco-friendly consumer choices by using Cleanyst's plant-based products to reduce plastic waste and carbon emissions.",
    "Empowering communities and individuals through education and involvement in sustainable practices to combat climate change effectively.",
    "Embracing sustainable living through eco-friendly consumer choices, corporate sustainability efforts, and supporting green energy and conservation initiatives.",
    "Choosing sustainable, skin-friendly products like bamboo bandages as a simple yet effective way to support eco-friendly consumer choices.",
    "Supporting local leadership and community initiatives that promote sustainability, equitable growth, and eco-friendly consumer practices in urban development." \\
    \hline
    PlasticWste\&EnvImpact & "Recognizing plastic's sustainability benefits, urging scientific studies and informed policies to optimize its use and minimize environmental impact.",
    "Reducing plastic waste through reusable products, supporting cleanup efforts, and embracing innovative solutions to protect marine ecosystems.",
    "Using pyrolysis technology to recycle waste plastic, supporting a circular economy and environmental protection efforts.",
    "Urgent collective action, habit changes, and support for organizations fighting plastic pollution to protect oceans and marine life.",
    "Eengaging brands towards Plastic Neutrality and supports innovative recycling technologies to combat the global plastic waste crisis.",
    "Daily action against plastic waste, supporting legislation, and making conscious choices in food consumption and waste management.",
    "Responsible waste management, recycling efforts, and addressing illegal activities to mitigate environmental contamination and promote sustainability." \\
    \hline
    PromtSustainTrnsport & "Embracing sustainable transportation and clean energy through EV test drives and educational events on innovative energy solutions.",
    "Attending Drive Electric Earth Day events to learn about and embrace the benefits of electric vehicles for a sustainable future.",
    "Adopting cycling and supporting policies for electric school buses and a bold climate bill for sustainable transportation.",
    "Urgent investment in electric vehicles and legislative support to reduce emissions and create union jobs.",
    "Adopting electric vehicles for their environmental benefits, cost savings, and the expanding availability of models.",
    "Adopting clean car standards and investing in EV infrastructure to improve air quality, health, and combat climate change.",
    "Significant investment in public transportation electrification and infrastructure to support climate goals, community benefits, and union job creation.",
    "Adopting and promoting sustainable transportation solutions and technologies to improve air quality and address environmental challenges.",
    "Infrastructure investments and community initiatives that reduce emissions and promote sustainable transportation as essential to combating climate change.",
    "Adopting biking, walking, electric vehicles, and public transit as key strategies to reduce emissions and improve air quality.",
    "Shifting to public and active transportation, electric vehicles, and renewable energy to reduce emissions and enhance sustainability." \\
    WaterMngmnt\&Sustain &  "Stricter regulations on coal ash disposal to protect water quality and public health in Georgia.",
    "Supporting legislative efforts to protect and restore Colorado's rivers for economic growth, climate resilience, and agricultural preservation.",
    "Exploring and adopting successful global strategies to address water scarcity and ensure sustainable water management in rapidly growing states.",
    "Sustainable water management through refurbishing existing facilities and adopting eco-friendly water solutions to protect health and the environment.",
    "Proactive water management strategies, infrastructure investment, and community involvement to ensure sustainable water access and conservation efforts.",
    "Responsible disposal of hazardous wastes like oils and grease to protect water sources and promote environmental sustainability.",
    "Innovative financing, policy reforms, and sustainable practices to ensure water security and resilience against climate change impacts.",
    "Active engagement and informed leadership in water board governance to ensure sustainable water management and protection." \\
\hline
\end{longtable}

\begin{longtable}{p{3.4cm}|p{9.8cm}}
\caption{Resulting arguments related to COVID-19 vaccine campaigns after two rounds of iterations.} \label{arg_covid} \\
\hline
Themes & Talking Points \\
\hline
\endfirsthead

\multicolumn{2}{c}%
{{\bfseries Table \thetable\ continued from previous page}} \\
\hline
Themes & Talking Points \\
\hline
\endhead

\hline \multicolumn{2}{r}{{Continued on next page}} \\ 
\endfoot

\hline
\endlastfoot

GovDistrust & "Accountability and reform in the Biden administration, criticizing its policies for undermining national stability and public trust.",
    "Against government control of the prescription drug market, warning of hindered medical innovation and reduced access to cures.",
    "Skepticism towards government actions and legislation, emphasizing the need for transparency and resistance to perceived overreach.",
    "Skepticism towards the Biden administration's handling of COVID-19, immigration, and economic policies.",
    "Questions the efficacy and motives of government policies on immigration, economy, and public health measures." \\
\hline
GovTrust & "Recognizing and appreciating government efforts and collaboration in pandemic response, emphasizing trust in legislative support for public health.",
    "Recognizing the Biden administration's efforts to address key national and global issues, building trust through proactive policies and actions.",
    "Political engagement and trust through transparency, community involvement, and proactive governance initiatives.",
    "Building trust through community engagement, open dialogue, and collaboration between citizens and their representatives.",
    "President Biden's commitment to international accords, public health, economic recovery, and progressive policy achievements to build government trust." \\
    \hline
    VaccineRollout & "The widespread and accessible COVID-19 vaccination and testing services, emphasizing no-appointment-necessary clinics to increase public participation.",
    "Utilizing digital platforms to access up-to-date information on COVID-19 vaccine availability and registration across different locations.",
    "Easy and accessible COVID-19 vaccination for all eligible individuals, emphasizing convenience and inclusivity in public health efforts.",
    "The organized distribution of COVID-19 vaccines across Texas, highlighting state efforts to enhance accessibility and efficiency in vaccination.",
    "The importance of local news in keeping communities informed about the evolving COVID-19 vaccine rollout and registration processes.",
    "Accessible vaccine clinics, incentive programs, and safety protocols to enhance vaccine rollout at the local level."\\
    \hline
    VaccineSymptom & "Emphasizes the safety and effectiveness of COVID-19 vaccines despite mild side effects, advocating for vaccination to protect against the virus.",
    "Vigilance and thorough investigation into the rare but serious side effects like blood clots associated with the J\&J vaccine.",
    "Careful monitoring and transparent communication about potential vaccine side effects to ensure public safety and confidence.",
    "Informed vaccine choices, awareness of rare side effects, and the importance of continued vaccination for public health safety.",
    "Awareness and transparency regarding the range and prevalence of COVID-19 vaccine-related symptoms and effects.",
    "Advocates for booster shots to enhance immunity, addressing vaccine safety and mild side effects experienced post-vaccination.",
    "Advocates for the safety of COVID-19 vaccines for children, emphasizing mild side effects and community protection through vaccination." \\
    \hline
    VaccineEquity & "Against price control legislation, emphasizing its potential to hinder vaccine development and equitable access to life-saving COVID-19 vaccinations.",
    "Global vaccine equity, emphasizing free and universal access to COVID-19 vaccines to end the pandemic for all.",
    "Overcoming barriers to ensure equitable COVID-19 vaccine access for underserved communities and prioritizing vulnerable populations.",
    "Investigating and improving the distribution and administration process to ensure equitable vaccine access in Illinois.",
    "Equitable access to COVID-19 vaccines, addressing eligibility, distribution fairness.",
    "Donating funds that will be matched to multiply impact, promoting fair global access to vaccines.",
    "Efforts and challenges in equitable vaccine distribution and access for seniors across various counties." \\
    \hline
    VaccineStatus & "Raising awareness about the low vaccination rates in Central Texas counties, as indicated by state data.",
    "Staying informed about COVID-19 vaccination updates and initiatives through trusted local news apps in respective counties.",
    "Accessing COVID-19 vaccine information and locations for vaccination through the SmartNews app for free.",
    "Highlights the concern over the high number of unvaccinated individuals in Central Texas counties according to state data." \\
    \hline
    EncourageVaccination & "Promoting vaccination by providing information, guidance, and incentives to encourage people to get vaccinated against COVID-19.",
    "Importance of COVID-19 vaccination by emphasizing its safety, efficacy, availability, and role in ending the pandemic.",
    "Engaging in community discussions and raising awareness about vaccination, especially within the LatinX community.",
    "Encourages COVID-19 vaccination and boosters for all ages to protect against the virus and maintain community health.",
    "Promoting COVID-19 vaccination for children and the community, emphasizing safety, accessibility, and the benefits of vaccination." \\
    \hline
    VaccineMandate & "Against COVID vaccine mandates and passports, urging support for personal choice and constitutional rights regarding vaccination.",
    "Awareness and updates on issues related to COVID-19 vaccine mandates and refusals in various states.",
    "COVID-19 vaccine mandates, including recent Supreme Court rulings and public opinion on federal mandates.",
    "Against Biden's vaccine mandate, portraying it as government tyranny and urging resistance and financial support.",
    "Advocating for Covid-19 vaccine mandates for customers by Snowy Range Taxi in Wyoming to prioritize community safety.",
    "Awareness and preparation among employers for the impending COVID-19 vaccine mandate.",
    "Against federal COVID-19 vaccine and mask mandates, framing them as overreach and questioning their constitutionality.",
    "Resistance against federal vaccine mandates, emphasizing personal choice and legal action to uphold individual rights.",
    "Vaccine documentation management, vaccine refusal, job consequences for non-compliance, and HIPAA implications in vaccine proof requests.",
    "Advocates for personal choice and against federal vaccine mandates, viewing them as infringements on civil liberties and rights." \\
    \hline
    VaccineReligion & "Using religion as a means to exempt individuals from mandatory COVID-19 vaccination requirements.",
    "Incorporating religious teachings into actions promoting community well-being, including vaccination and pandemic safety measures.",
    "Vaccination as an act of love and protection towards neighbors, aligning it with Christian values.",
    "Seeking religious guidance and perspective on COVID, vaccination, and related moral concerns with Pastor Bunjee Garrett.",
    "Religious communities to participate in public health by supporting vaccination as an act of neighborly love and care." \\
    \hline
    VaccineEfficacy & "The safety and efficacy of the Moderna COVID-19 vaccine, but acknowledges instances of post-vaccination infections.",
    "Getting vaccinated, highlighting the vaccine's effectiveness in significantly reducing infection, severity, and mortality from the Delta variant of COVID-19.",
    "Strongly advocates for COVID-19 vaccination, highlighting its safety, efficacy, and crucial role in preventing severe illness and ending the pandemic.",
    "The prioritization strategy for COVID-19 vaccinations in certain highly infected North Carolina counties, addressing vaccine efficacy concerns.",
    "Promoting vaccination as an effective measure to protect individuals and families from COVID-19, with community endorsement.",
    "Highlights that vaccinations, including boosters, are highly effective in preventing severe illness, hospitalization, and death from COVID-19.",
    "Advocates for vaccination and continued safety measures to protect against the transmissible and potentially severe Delta variant." \\
    \hline
    VaccineDevelopment & "Trust in COVID-19 vaccines, highlighting their development from decades-old, scientifically robust messenger RNA technology.",
    "Confidence in the Pfizer-BioNTech COVID-19 vaccine, emphasizing its emergency use authorization and pending full FDA approval.",
    "Understanding the distinct characteristics and allocation strategies of the newly approved Johnson \& Johnson COVID-19 vaccine." \\
    \hline
    CovidPlan & "Staying informed about regional COVID-19 vaccine distribution plans and understanding challenges in meeting demand.",
   "Recognizing and supporting the American Rescue Plan and lawmakers' efforts in providing pandemic relief.",
   "Investigating and addressing the apparent inefficiencies in the COVID-19 vaccine distribution in Illinois.",
   "Commending the collaborative development of COVID-19 vaccines and treatments supported by the American Rescue Plan.",
   "Providing COVID-19 vaccinations to residents through the \#VaxMobile program in different locations throughout the township." \\
   \hline
   VaccineMisinformation & "Correcting vaccine misinformation and encouraging vaccine participation through expert-led educational conversations.",
    "Debunking vaccine myths and empowering people with credible and accurate information about COVID-19 vaccines.",
    "Holding news outlets accountable for spreading potentially harmful misinformation about Covid-19 vaccines.",
    "Clear communication and spreading accurate information about COVID-19 vaccines to counter misinformation.",
    "Utilizing virtual town hall meetings to combat vaccine misinformation by providing accurate, expert information. \\
    \hline
    NaturalImmunity & "Changes in herd immunity goals due to the Delta variant.",
    "Natural immunity against COVID-19 could be better than vaccine-induced immunity",
    "Boosting the immune system naturally with supplements as an alternative approach to handling COVID-19." \\
    \hline
\end{longtable}
\twocolumn

%
\end{document}